\documentclass[11pt,a4paper]{article}
\usepackage{CJK}
\usepackage{hyperref}
\usepackage{acl2017}
\usepackage{times}
\usepackage{latexsym}

\usepackage{CJK}
\usepackage{graphicx}
\usepackage{amsmath}
\usepackage{amsfonts}
\usepackage{bm}
\usepackage{mathrsfs}
\usepackage{paralist}
\usepackage{graphicx} 
\usepackage{float}
\usepackage{subfigure}
\usepackage{booktabs,multirow}
\usepackage{bigstrut,bigdelim}

\definecolor{ruber}{rgb}{0.80469,0.27344,0.46094}

\aclfinalcopy 
\def\aclpaperid{***} 

\newcommand{\ruber}{\textsc{Ruber}}
\newcommand{\bleu}{\textsc{Bleu}}
\newcommand{\rouge}{\textsc{Rouge}}

\title{{\color{ruber}\ruber}: An Unsupervised Method for Automatic Evaluation\\ of Open-Domain Dialog Systems
}

\author{Chongyang Tao,$^1$\quad Lili Mou,$^2$\quad Dongyan Zhao,$^1$\quad Rui Yan$^1$\\
	$^1$Institute of Computer Science and Technology, Peking University, China\\
	$^2$Key Laboratory of High Confidence Software Technologies (Peking University)\\
	Ministry of Education, China; Institute of Software, Peking University, China\\
	\texttt{\{chongyangtao,zhaody,ruiyan\}@pku.edu.cn}\\
	\texttt{doublepower.mou@gmail.com}}

\date{}

\begin{document}
\begin{CJK*}{UTF8}{gkai}
\maketitle
\begin{abstract}
Open-domain human-computer conversation has been attracting increasing attention over the past few years.
However, there does not exist a standard automatic evaluation metric for open-domain dialog systems; researchers usually resort to human annotation for model evaluation, which is time- and labor-intensive.
In this paper, we propose \ruber, a \textit{Referenced metric and Unreferenced metric Blended Evaluation Routine}, 
which evaluates a reply by taking into consideration both a groundtruth reply and a query (previous user-issued utterance).
Our metric is learnable, but its training does not require labels of human satisfaction. Hence, \ruber\ is flexible and extensible to different datasets and languages. Experiments on both retrieval and generative dialog systems show that \ruber\ has a high correlation with human annotation.
\end{abstract}
\section{Introduction}

Automatic evaluation is crucial to the research of open-domain human-computer dialog systems. Nowadays, open-domain conversation is attracting increasing attention as an established scientific problem~\cite{bickmore2005establishing,bessho2012dialog,NRM}; it also has wide industrial applications like XiaoIce\footnote{\url{http://www.msxiaoice.com/}} from Microsoft and DuMi\footnote{\url{http://duer.baidu.com/}} from Baidu. Even in task-oriented dialog (e.g., hotel booking), an open-domain conversational system could be useful in handling unforeseen user utterances.

In existing studies, however, researchers typically resort to manual annotation to evaluate their models, which is expensive and time-consuming. Therefore, automatic evaluation metrics are particularly in need, so as to ease the burden of model comparison and to promote further research on this topic.

In early years, traditional vertical-domain dialog systems use metrics like slot-filling accuracy and goal-completion rate \cite{walker1997paradise,walker2001quantitative,schatzmann2005quantitative}. Unfortunately, such evaluation hardly applies to the open domain due to the diversity and uncertainty of utterances: ``accurancy" and ``completion," for example, make little sense in open-domain conversation.

Previous studies in several language generation tasks have developed successful automatic evaluation metrics, e.g., \textsc{Bleu}~\cite{papineni2002bleu} and \textsc{Meteor}~\cite{banerjee2005meteor} for machine translation, and \textsc{Rouge}~\cite{rouge} for summarization. For dialog systems, researchers occasionally adopt these metrics for evaluation~\cite{ritter2011data,li2015diversity}. 
However, \newcite{liu2016not} conduct extensive empirical experiments and show weak correlation between existing metrics and human annotation. 

Very recently, \newcite{lowe2017towards} propose a neural network-based metric for dialog systems; it learns to predict a score of a reply given its query (previous user-issued utterance) and  a groundtruth reply.
But such approach requires massive human-annotated scores to train the network, and thus is less flexible and extensible. 

In this paper, we propose \textsc{Ruber}, a \textit{Referenced metric and Unreferenced metric Blended Evaluation Routine} for open-domain dialog systems. 
\ruber\ has the following distinct features:
\begin{compactitem}
	\item An embedding-based scorer measures the similarity between a generated reply and the groundtruth.
	We call this a \textit{referenced} metric, because it uses the groundtruth as a reference, akin to existing evaluation metrics. Instead of using word-overlapping information (e.g., in \textsc{Bleu} and \textsc{Rouge}), we measure the similarity by pooling of word embeddings; it is more suited to dialog systems due to casual expressions in open-domain conversation.
	\item A neural network-based scorer measures the relatedness between the generated reply and its query.
	We observe that the query-reply relation is informative itself. This scorer is \textit{unreferenced} because it does not refer to groundtruth. We apply negative sampling to train the network. Our approach requires no manual annotation label, and hence is more extensible than~\newcite{lowe2017towards}.
	\item We propose to combine the referenced and unreferenced metrics to better make use both worlds. On the one hand, closeness to groundtruth implies high quality. On the other hand, the groundtruth does not cover all possible ways of replying the query; the relatedness between a generated reply and its query then provides additional information. Combining these two aspects with heuristic strategies (e.g., averaging) further improves the performance.
\end{compactitem}

In this way, \ruber\ does not require human annotation scores for training, in the sense of which, we call our metric \textit{unsupervised}. Although we still have to prepare a corpus to train embeddings (in an unsupervised manner) and neural scorers (by negative sampling), the query-reply data---also a prerequisite in \newcite{lowe2017towards}---are much cheaper to obtain than human annotation of their satisfaction, showing the advantage of our approach.

We evaluated \ruber\ on prevailing dialog systems, including both retrieval and generative ones. Experiments show that \ruber\  significantly outperforms existing automatic metrics in terms of the Pearson and Spearman correlation with human judgments, and has fair transferability over different open-domain datasets.

\section{Empirical Observations}

\begin{table}[!t]
	\centering
	\resizebox{\linewidth}{!}{
		\begin{tabular}{rl}
			\hline
			\textbf{Query}:\!\!&\!\!为什么不领养一个？\\
			\!\!&\!\!Why not adopt one?\\
			\textbf{Groundtruth}:\!\!&\!\!我也想领养，但是不好找\\
			\!\!&\!\!I hope so, but it's hard to find one\\
			\midrule
			\textbf{Candidate 1}:\!\!&\!\!恩，我已经在朋友中找了\\
			\!\!&\!\!I'm looking for one from my friends\\
			\textbf{Candidate 2}:\!\!&\!\!巴比伦是稀有品种\\
			\!\!&\!\!Pappilon is a rare breed\\
			\textbf{Candidate 3}:\!\!&\!\!可以哈，谢谢你的建议\\
			\!\!&\!\!OK, thank you for your advice\\
			\hline
		\end{tabular}
	}
	\caption{Query and groundtruth/candidate replies.}\label{tab:example}
\end{table}
\begin{figure}[!t]
	\centering
	\includegraphics[width=.9\linewidth]{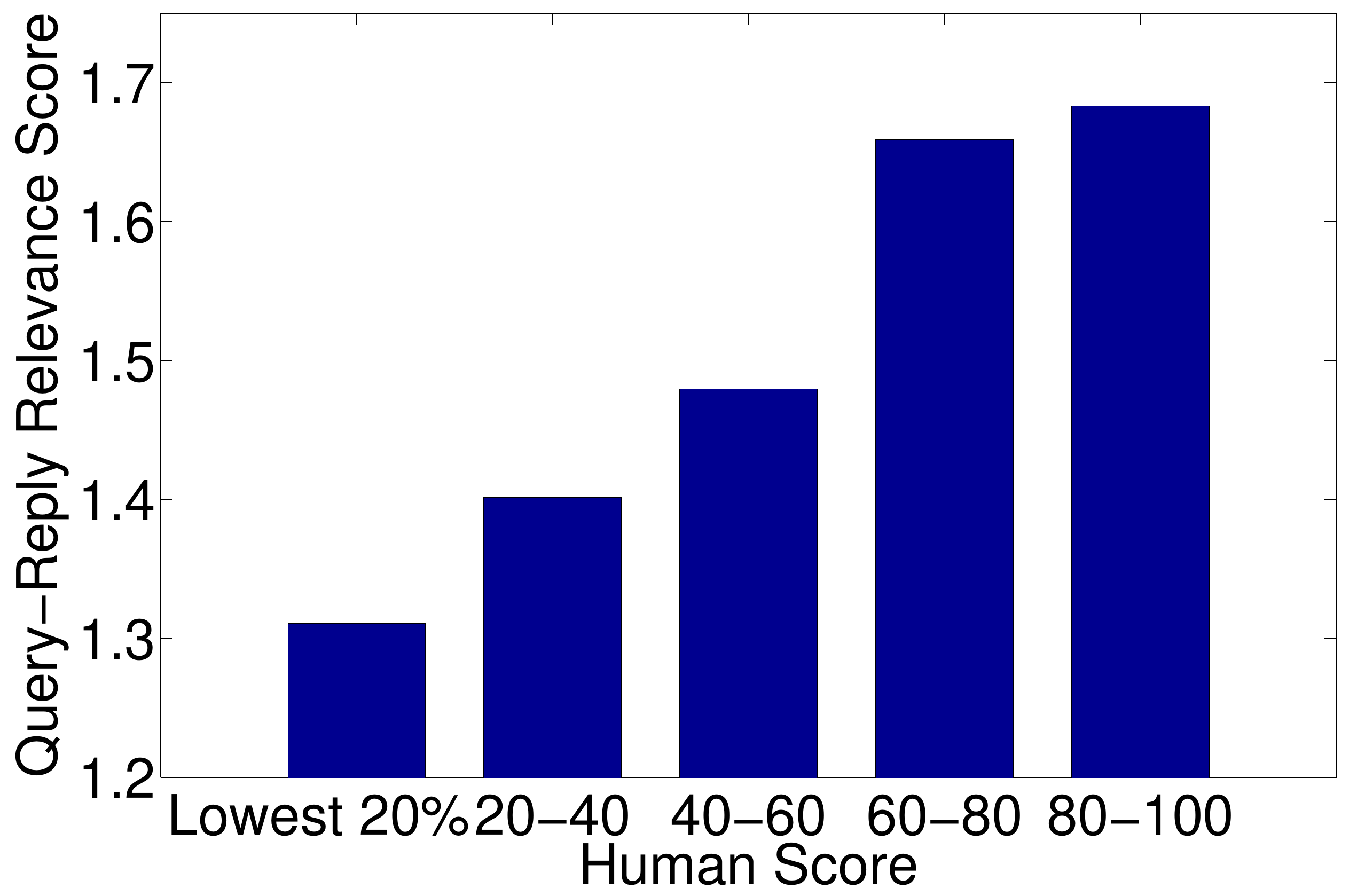}
	\caption{Average query-reply relevance scores versus quantiles of human scores. In other words,
		we divide human scores (averaged over all annotators) into 5 equal-sized groups, and show the average
		query-reply relevance score (introduced in Section~\ref{ss:ref}) of each group.}\label{fig:qr}
\end{figure}
In this section, we present our empirical observations regarding the question ``\textit{What makes a good reply in open-domain dialog systems?}'' 

\textbf{Observation 1.} Resembling the groundtruth generally implies a good reply. This is a widely adopted assumption in almost all metrics, e.g., \bleu, \rouge, and \textsc{Meteor}.
However, utterances are typically short and casual in dialog systems; thus word-overlapping statistics are of high variance. Candidate 1 in Table~\ref{tab:example}, for example, resembles the groundtruth in meaning, but shares only a few common words. Hence our method measures similarity based on embeddings.

\textbf{Observation 2.} A groundtruth reply is merely one way to respond. Candidate 2 in Table~\ref{tab:example} illustrates a reply that is different from the groundtruth in meaning but still remains a good reply to the query. Moreover, a groundtruth reply may be universal itself (and thus undesirable). ``I don't know,''---which appears frequently in the training set~\cite{li2015diversity}---may also fit the query, but it does not make much sense in a commercial chatbot.\footnote{Even if a system wants to mimic the tone of humans by saying ``I don't know,'' it can be easily handled by post-processing. The evaluation then requires system-level information, which is beyond the scope of this paper.} The observation implies that a groundtruth alone is insufficient for the evaluation of open-domain dialog systems.

\textbf{Observation 3.}  Fortunately, a query itself provides useful information in judging the quality of a reply.\footnote{Technically speaking, a dialog generator is also aware of the query. However, a discriminative model (scoring a query-reply pair) is more easy to train than a generative model (synthesizing a reply based on a query). There could also be possibilities of generative adversarial training.} Figure~\ref{fig:qr} plots the average human satisfactory score of a groundtruth reply versus the relevance measure (introduced in Section~\ref{ss:unref}) between the reply and its query. We see that, even for groundtruth replies, those more relevant to the query achieve higher human scores.
The observation provides rationales of using query-reply information as an unreferenced score in dialog systems.

\begin{figure}[!t]
	\centering
	\includegraphics[width=\linewidth]{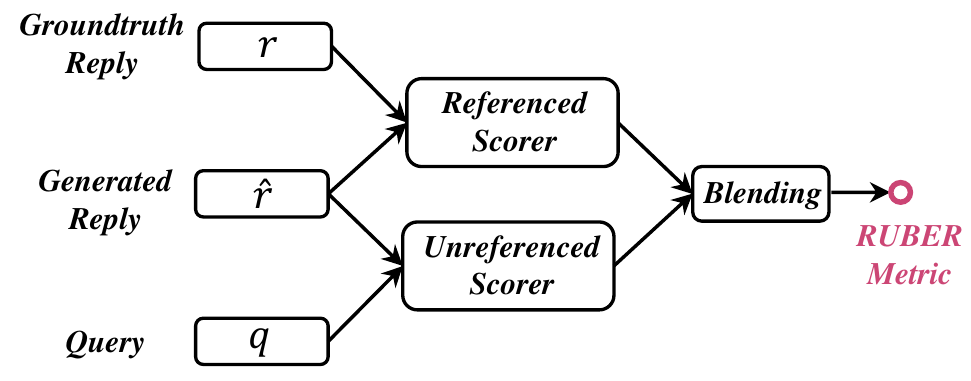}
	\caption{Overview of the \ruber\ metric.}
	\label{fig:ruber}
\end{figure}

\section{Methodology}\label{sec:method}

Based on the above observations, we design referenced and unreferenced metrics in Subsections~\ref{ss:ref} and~\ref{ss:unref}, respectively; Subsection~\ref{ss:combine} discusses how they are combined. The overall design methodology of our \ruber\ metric is also shown in Figure~\ref{fig:ruber}.

\subsection{Referenced Metric}\label{ss:ref}
We measure the similarity between a generated reply $\hat{r}$ and a groundtruth $r$ as a referenced metric.
Traditional referenced metrics typically use word-overlapping information including both precision (e.g., \textsc{Bleu}) and recall (e.g., \textsc{Rouge})~\cite{liu2016not}.
As said, they may not be appropriate for open-domain dialog systems.

We adopt the vector pooling approach that summarizes sentence information by choosing the maximum and minimum values in each dimension; the closeness of a sentence pair is measured by the cosine score.
We use such heuristic matching because we assume no groundtruth scores, making it infeasible to train a parametric model.

Formally, let $\bm w_1, \bm w_2, \cdots, \bm w_n$ be the embeddings of words in a sentence, max-pooling summarizes the maximum value as
\begin{equation}
	\bm v_{\max}[i]=\max\big\{\bm w_1[i], \bm w_2[i], \cdots, \bm w_n[i]\big\}
\end{equation}
where $[\cdot]$ indexes a dimension of a vector. Likewise, min pooling yields a vector $\bm v_{\min}$.
Because an embedding feature is symmetric in terms of its sign, we concatenate both max- and min-pooling vectors as $\bm v = [\bm{ v}_{\max};\bm v_{\min}]$. 

Let ${\bm v}_{\hat r}$ be the generated reply's sentence vector and $\bm v_{r}$ be that of the groundtruth reply, both obtained by max and min pooling. The referenced metric $s_R$ measures the similarity between $r$ and $\hat r$ by

\begin{equation}
	s_R(r,\hat{r}) = \cos(\bm v_r, {\bm v}_{\hat{r}})=\frac{\bm v_{r}^\top{\bm v}_{\hat r}}{\|\bm v_r\|\cdot\|{\bm v}_{\hat r}\|}
\end{equation}

\newcite{forgues2014bootstrapping} propose a vector extrema method that utilizes embeddings by choosing either the largest positive or smallest negative value. Our heuristic here is more robust in terms of the sign of a feature.

\subsection{Unreferenced Metric}\label{ss:unref}
\begin{figure}[!t]
	\centering
	\includegraphics[width=\linewidth]{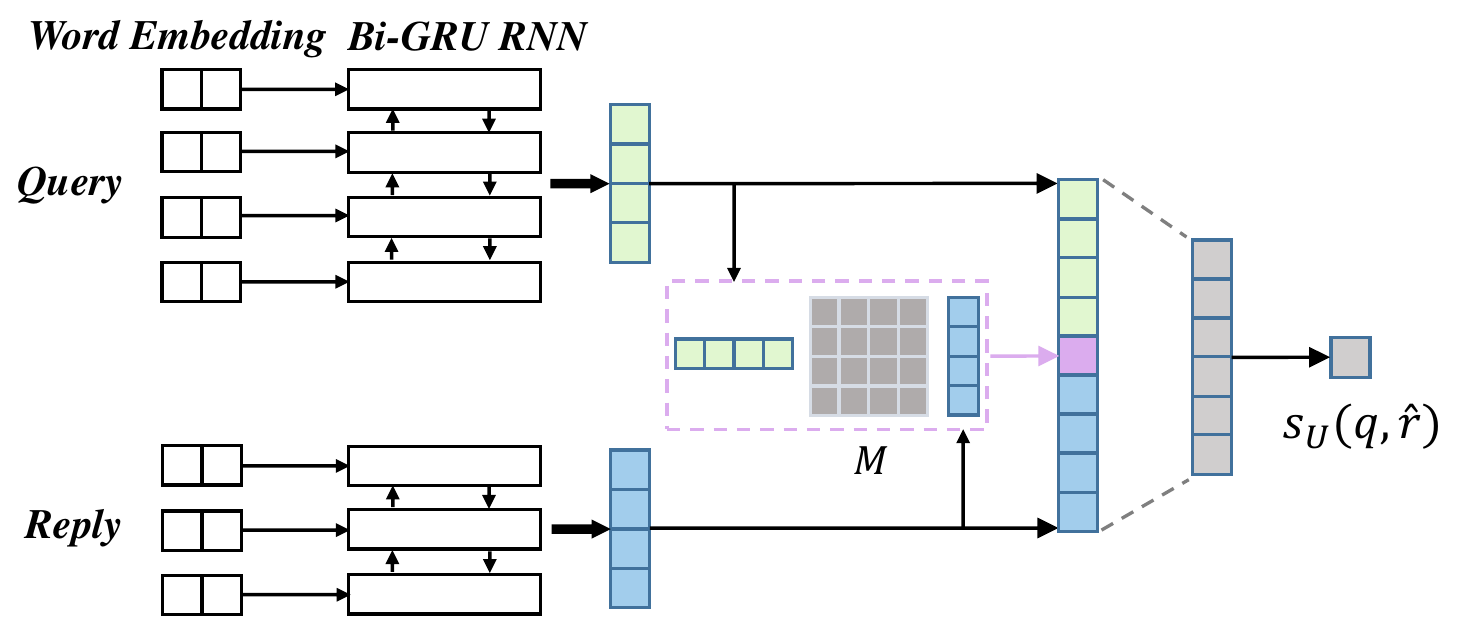} 
	\caption{The neural network predicting the unreferenced score.}
	\label{fig:nn}
\end{figure}
We then measure the relatedness between the generated reply $\hat r$ and its query $q$. This metric is unreferenced and denoted as $s_U(q, \hat r)$, because it does not refer to a groundtruth reply.

Different from the $r$-$\hat r$ metric, which mainly measures the similarity of two utterances, the $q$-$\hat r$ metric in this part involves more semantics. Hence, we empirically design a neural network (Figure~\ref{fig:nn}) to predict the appropriateness of a reply with respect to a query.

Concretely, each word in a query $q$ and a reply $r$ is mapped to an embedding; a bidirectional recurrent neural network with gated recurrent units (Bi-GRU RNN) captures information along the word sequence. The forward RNN takes the form
\begin{align}\nonumber
	\bm [\bm r_t; \bm z_t]&= \sigma(W_{r,z}{\bm x}_t + U_{r,z}\bm h_{t-1}^\rightarrow+\bm b_{r,z})\\
	\nonumber
	\tilde{\bm h}_t&=\tanh\big(W_h{\bm x}_t+U_h(\bm r_t\circ\bm h_{t-1}^\rightarrow)+\bm b_h\big)\\\nonumber
	\bm h_t^\rightarrow&=(1-\bm z_t)\circ \bm h_{t-1}^\rightarrow+\bm z_t\circ\tilde{\bm h}_t
\end{align}
where $\bm x_t$ is the embedding of the current input word, and $\bm h_t^\rightarrow$ is the hidden state. Likewise, the backward RNN gives hidden states $\bm h_t^\leftarrow$. 
The last states of both directions are concatenated as the sentence embedding ($\bm q$ for a query and $\bm r$ for a reply).

We further concatenate $\bm q$ and $\bm r$ to match the two utterances. Besides, we also include a ``quadratic feature'' as $\bm q^\top\mathbf{M} \bm r$, where $\mathbf M$ is a parameter matrix.
Finally, a multi-layer perceptron (MLP) predicts a scalar score as our unreferenced metric $s_U$. The hidden layer of MLP uses $\tanh$ as the activation function, whereas the last (scalar) unit uses $\operatorname{sigmoid}$ because we hope the score is bounded.

The above empirical structure is mainly inspired by several previous studies~\cite{sigir15,yan2016learning}. We may also apply other variants for utterance matching~\cite{match,tbcnnpair}; details are beyond the focus of this paper. 

To train the neural network, we adopt negative sampling, which does not require human-labeled data. That is, given a groundtruth query-reply pair, we randomly choose another reply $r^-$ in the training set as a negative sample. We would like the score of a positive sample to be larger than that of a negative sample by at least a margin $\Delta$. The training objective is to minimize
\begin{equation}
	J = \max\big\{0, \Delta - s_U(q, r) + s_U(q,r^-)\big\}
\end{equation}
All parameters are trained by \textit{Adam}~\cite{kingma2014adam} with backpropagation.

In previous work, researchers adopt negative sampling for utterance matching~\cite{yan2016learning}. Our study further verifies that negative sampling is useful for the evaluation task, which eases the burden of human annotation compared with fully supervised approaches that requirer manual labels for training their metrics~\cite{lowe2017towards}.

\subsection{Hybrid Approach}\label{ss:combine}

We combine the above two metrics by simple heuristics, resulting in a hybrid method \ruber\ for the evaluation of open-domain dialog systems.

We first normalize each metric to the range $(0, 1)$, so that they are generally of the same scale. In particular, the normalization is given by 
\begin{equation}
	\tilde s = \dfrac{ s- \min(s')}{\max(s')-\min(s')}
\end{equation}
where $\min(s')$ and $\max(s')$ refer to the maximum and minimum values, respectively, of a particular metric.

Then we combine $\tilde s_R$ and $\tilde s_U$ as our ultimate \ruber\ metric by heuristics including $\min$, $\max$, geometric averaging, and arithmetic averaging. As we shall see in Section~\ref{sec:quantitative}, different strategies yield similar results, consistently outperforming baselines. 

To sum up, \ruber\ metric is simple, general (without sophisticated model designs), and rather effective.
\section{Experiments}
\begin{table*}[htbp]
	\centering
	\resizebox{0.9\textwidth}{!}{
		\begin{tabular}{c|c|c|c|c|c}
			\toprule[1.5pt] \hline
			\multicolumn{2}{c|}{\multirow{2}[4]{*}{Metrics}}
			& \multicolumn{2}{c|}{\textbf{Retrieval (Top-1)}}
			& \multicolumn{2}{c}{\textbf{Seq2Seq (w/ attention) }} \\
			\cline{3-6}    \multicolumn{2}{c|}{} & Pearson\scriptsize($p$-value)  & Spearman\scriptsize($p$-value)  & Pearson\scriptsize(p-value)  & Spearman\scriptsize(p-value)  \\
			\hline  
			\hline
			\multirow{2}[2]{*}{Inter-annotator}
			& Human (Avg) & 0.4927{\scriptsize($<\!0.01$)} & 0.4981{\scriptsize($<\!0.01$)} & 0.4692{\scriptsize($<\!0.01$)} & 0.4708{\scriptsize($<\!0.01$)} \\
			& Human (Max) & 0.5931{\scriptsize($<\!0.01$)} & 0.5926{\scriptsize($<\!0.01$)} & 0.6068{\scriptsize($<\!0.01$)} & 0.6028{\scriptsize($<\!0.01$)} \\
			\hline  
			
			\multirow{6}[2]{*}{Referenced} & \bleu-1 & 0.2722{\scriptsize($<\!0.01$)} & 0.2473{\scriptsize($<\!0.01$)} & 0.1521\scriptsize($<\!0.01$) & 0.2358{\scriptsize($<\!0.01$)} \\
			& \bleu-2 & 0.2243{\scriptsize($<\!0.01$)} & 0.2389{\scriptsize($<\!0.01$)} &\!\!\!\,-0.0006{\scriptsize($0.9914$)} & 0.0546{\scriptsize($0.3464$)} \\
			& \bleu-3 & 0.2018{\scriptsize($<\!0.01$)} & 0.2247{\scriptsize($<\!0.01$)} & \!\!\!\,-0.0576{\scriptsize($0.3205$)} & \!\!\!\,-0.0188{\scriptsize($0.7454$)} \\
			& \bleu-4 & 0.1601{\scriptsize($<\!0.01$)} & 0.1719{\scriptsize($<\!0.01$)} & \!\!\!\,-0.0604{\scriptsize($0.2971$)} & \!\!\!\,-0.0539{\scriptsize($0.3522$)} \\
			& \rouge & 0.2840{\scriptsize($<\!0.01$)} & 0.2696{\scriptsize($<\!0.01$)} & 0.1747{\scriptsize($<\!0.01$)} & 0.2522{\scriptsize($<\!0.01$)} \\
			& Vector pool ($s_R$) & 0.2844{\scriptsize($<\!0.01$)} & 0.3205{\scriptsize($<\!0.01$)} & 0.3434\scriptsize($<\!0.01$) & 0.3219{\scriptsize($<\!0.01$)} \\
			\hline  
			
			\multirow{2}[2]{*}{Unreferenced} & Vector pool & 0.2253\scriptsize($<\!0.01$) & 0.2790\scriptsize($<\!0.01$) & 0.3808\scriptsize($<\!0.01$) &0.3584\scriptsize($<\!0.01$) \\
			& NN scorer ($s_U$)  & 0.4278\scriptsize($<\!0.01$) & 0.4338\scriptsize($<\!0.01$) & 0.4137\scriptsize($<\!0.01$) & 0.4240\scriptsize($<\!0.01$) \\
			\hline  
			
			\multirow{4}[2]{*}{\ruber} & Min   & 0.4428\scriptsize($<\!0.01$) & 0.4490\scriptsize($<\!0.01$) & \textbf{0.4527}\scriptsize($<\!0.01$) & \textbf{0.4523}\scriptsize($<\!0.01$) \\
			& Geometric mean & 0.4559\scriptsize($<\!0.01$) & 0.4771\scriptsize($<\!0.01$) & 0.4523\scriptsize($<\!0.01$) & 0.4490\scriptsize($<\!0.01$) \\
			& Arithmetic mean & \textbf{0.4594}\scriptsize($<\!0.01$) & \textbf{0.4906\scriptsize($<\!0.01$)} & 0.4509\scriptsize($<\!0.01$) & 0.4458\scriptsize($<\!0.01$) \\
			& Max   & 0.3263\scriptsize($<\!0.01$) & 0.3551\scriptsize($<\!0.01$) & 0.3868\scriptsize($<\!0.01$) & 0.3623\scriptsize($<\!0.01$) \\
			\hline
			\bottomrule[1.5pt]
		\end{tabular}
	}
	\caption{Correlation between automatic metrics and human annotation. The $p$-value is a rough estimation of the probability that an uncorrelated metric produces a result that is at least as extreme as the current one; it does not indicate the degree of correlation.}
	\label{tab:result}
\end{table*}

In this section, we evaluate the correlation between our \ruber\ metric and human annotation, which is the ultimate goal of automatic metrics. The experiment was conducted on a Chinese corpus because of cultural background, as human aspects are deeply involved in this paper. We also verify the performance of \ruber\ metric when it is transferred to different datasets. We believe our evaluation routine could be applied to different languages.

\subsection{Setup}\label{sec:set}
We crawled massive data from an online Chinese forum Douban.\footnote{\url{http://www.douban.com}} The training set contains 1,449,218 samples, each of which consists of a query-reply pair (in text). We performed Chinese word segmentation, and obtained Chinese terms as primitive tokens. In the referenced metric, we train 50-dimensional word2vec embeddings on the Douban dataset.

The \ruber\ metric (along with baselines) is evaluated on two prevailing dialog systems. One is a feature-based retrieval-and-reranking system, which first retrieves a coarse-grained candidate set by keyword matching and then reranks the candidates by human-engineered features; the top-ranked results are selected for evaluation~\cite{song2016two}. The other is a sequence-to-sequence (Seq2Seq) neural network~\cite{sutskever2014sequence} that encodes a query as a vector with an RNN and decodes the vector to a reply with another RNN; the attention mechanism~\cite{attention} is also applied to enhance query-reply interaction.

We had 9 volunteers to express their human satisfaction of a reply (either retrieved or generated) to a query by rating an integer score among 0, 1, and~2. A score of 2 indicates a ``good'' reply, 0 a bad reply, and 1 borderline.

\begin{figure*}[!t] \centering
	\subfigure[Human \scriptsize(1 vs. rest)] { \label{fig:top12}     
		\includegraphics[width=0.23\linewidth]{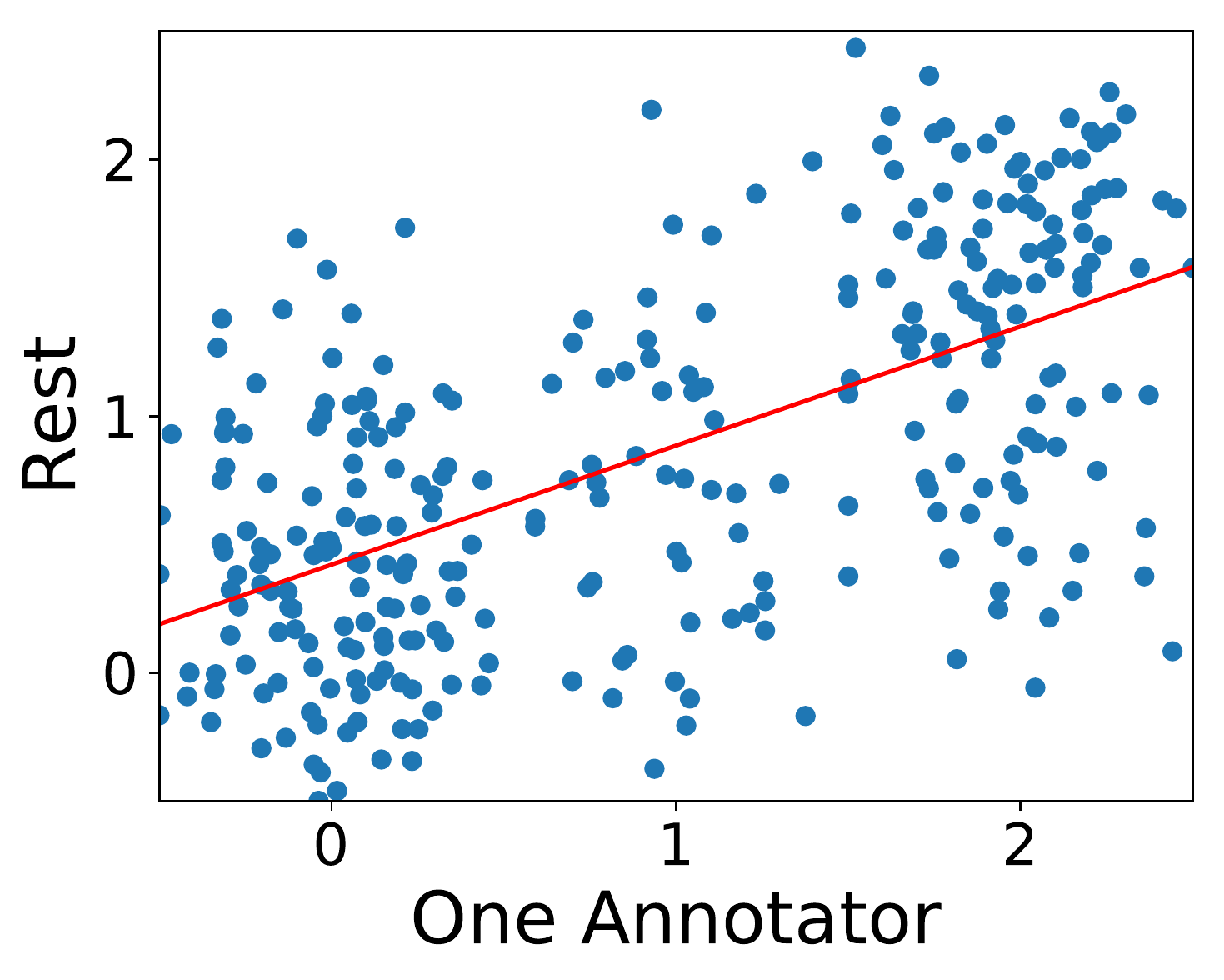} 
	}   
	\subfigure[Human \scriptsize(Group 1 vs. Group 2)] { \label{fig:top11}     
		\includegraphics[width=0.23\linewidth]{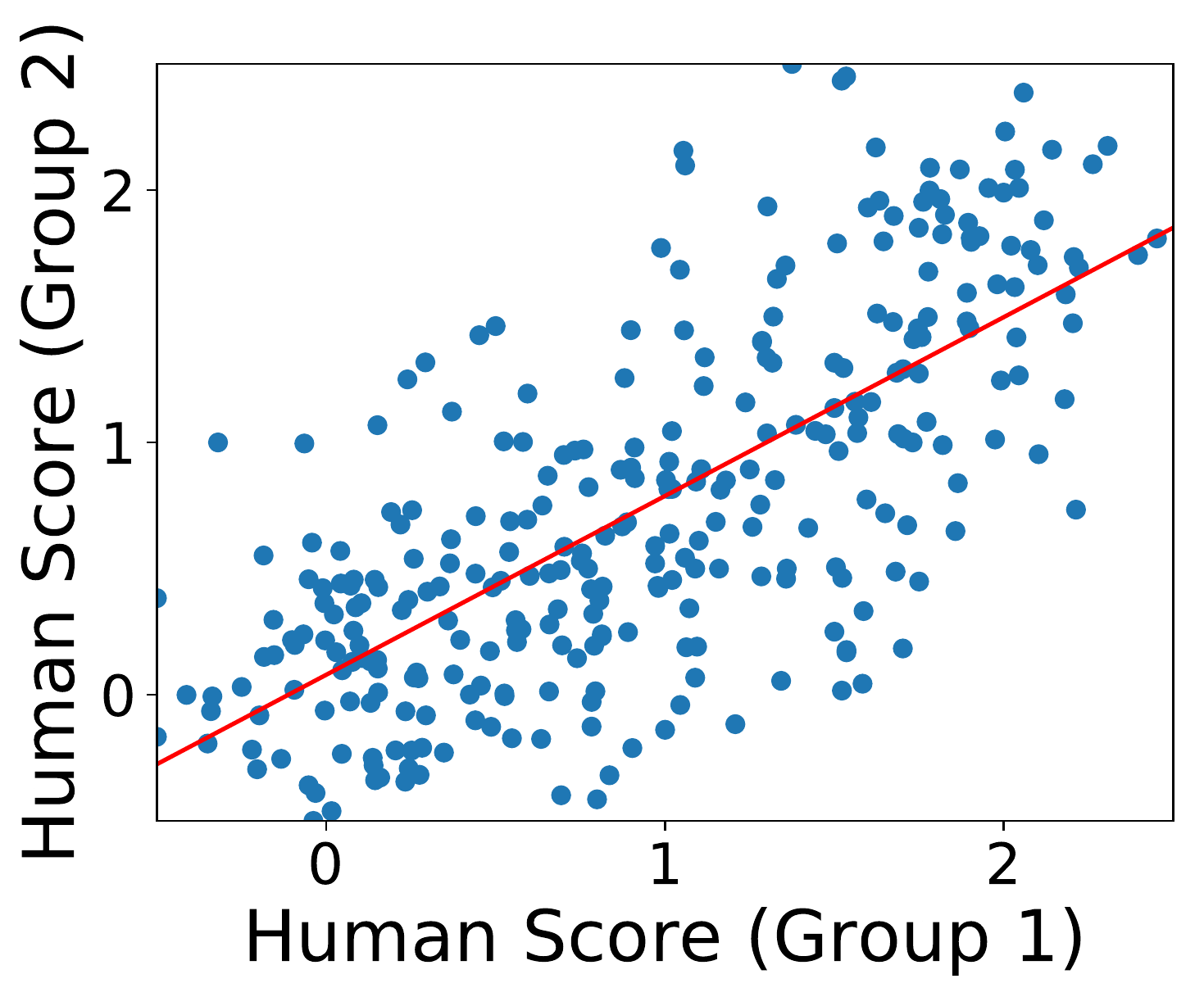} 
	}      
	\subfigure[\bleu-2] { \label{fig:top13}     
		\includegraphics[width=0.23\linewidth]{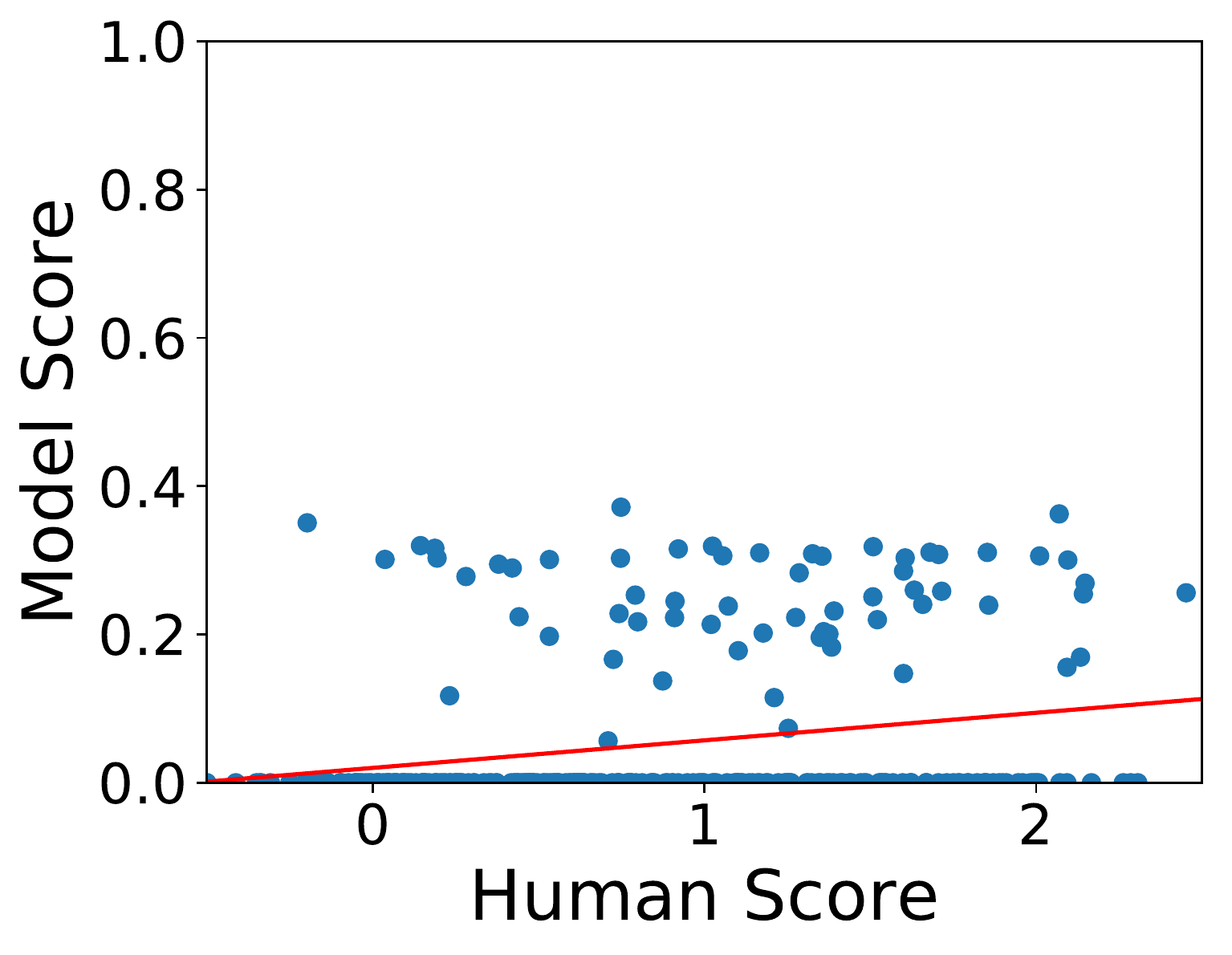} 
	}     
	\subfigure[\rouge] { \label{fig:top14}     
		\includegraphics[width=0.23\linewidth]{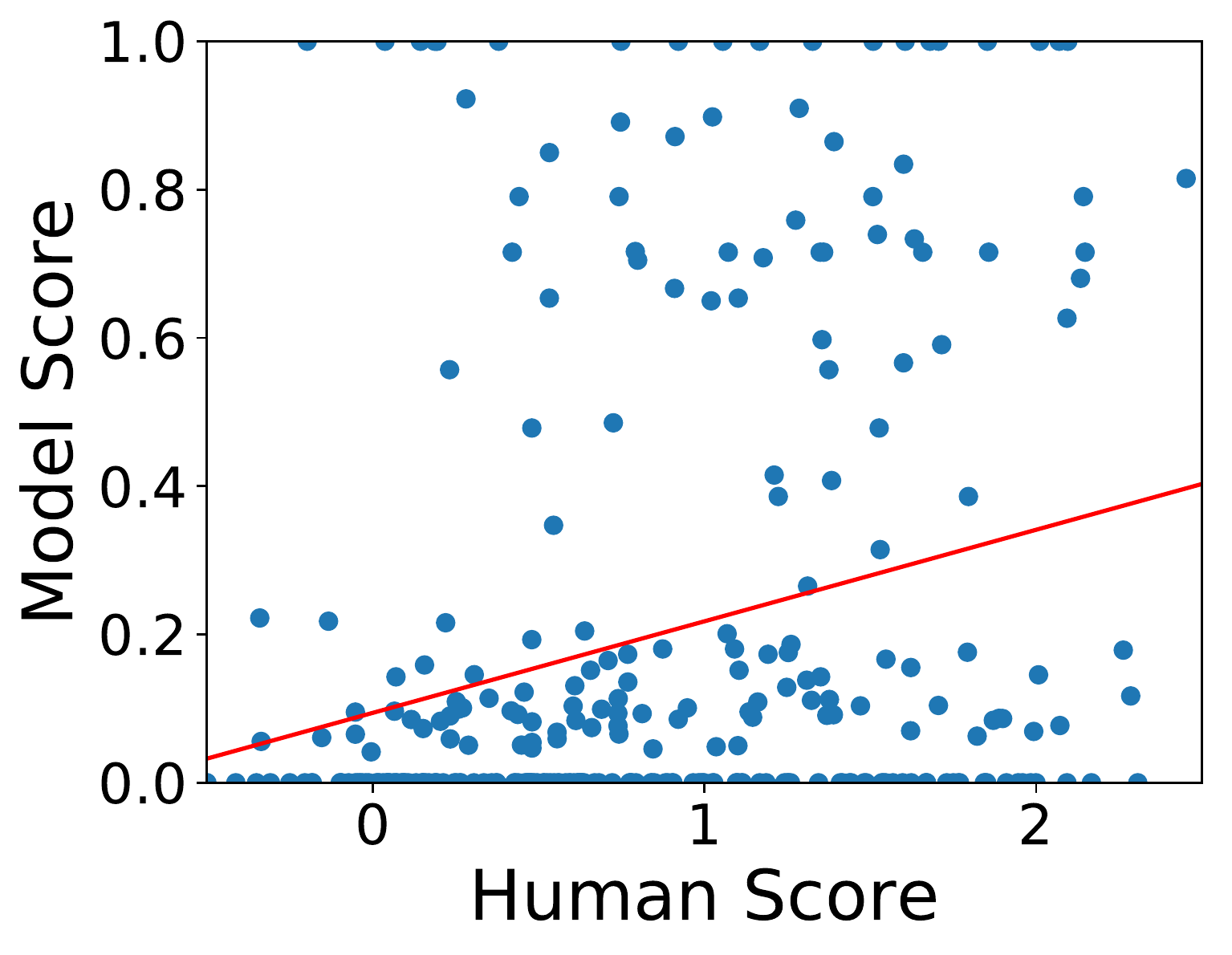} 
	}
	\subfigure[$s_R$ (vector pool)] { \label{fig:top15}     
		\includegraphics[width=0.23\linewidth]{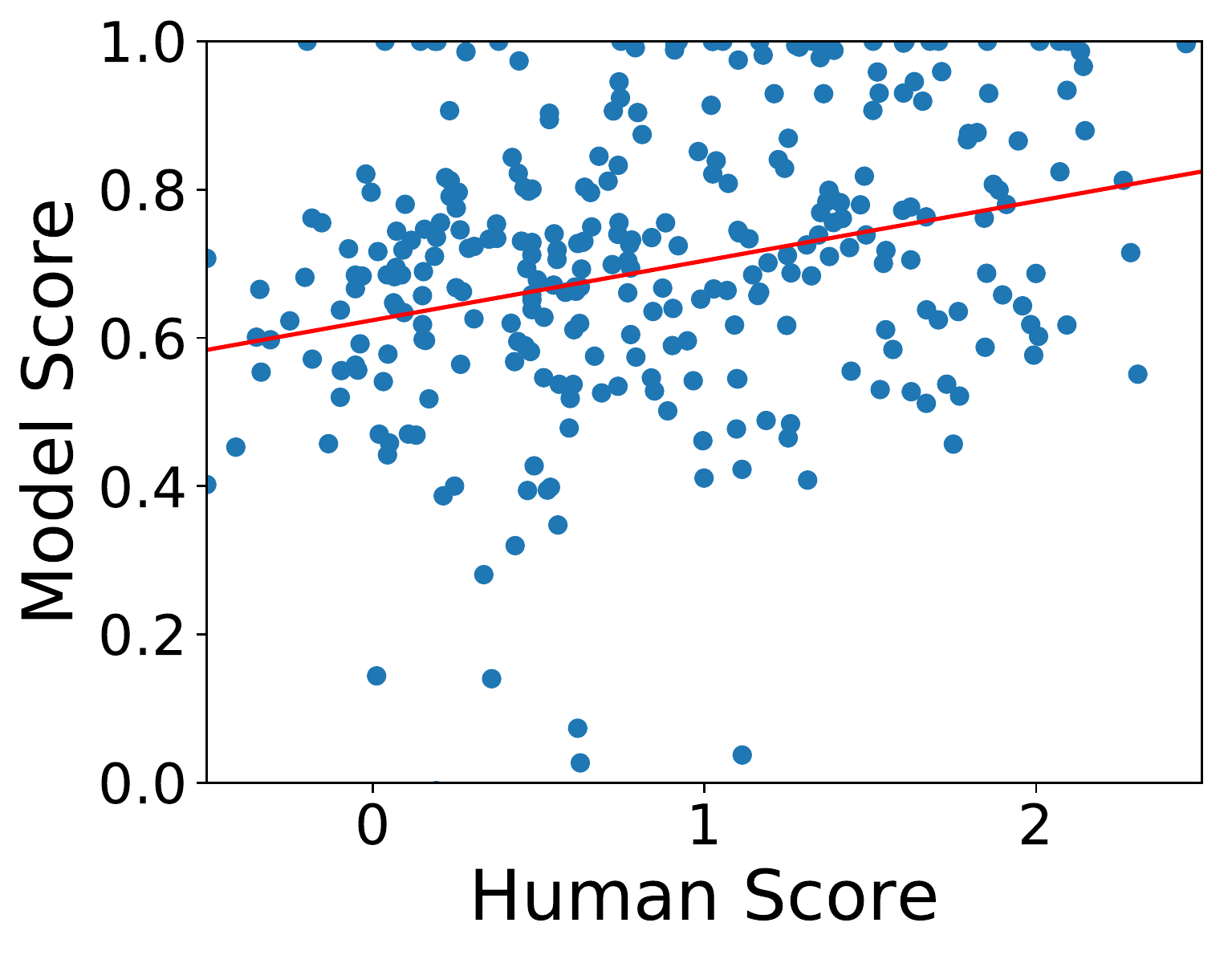} 
	}   
	\subfigure[$s_U$ (NN scorer)] { \label{fig:top16}     
		\includegraphics[width=0.23\linewidth]{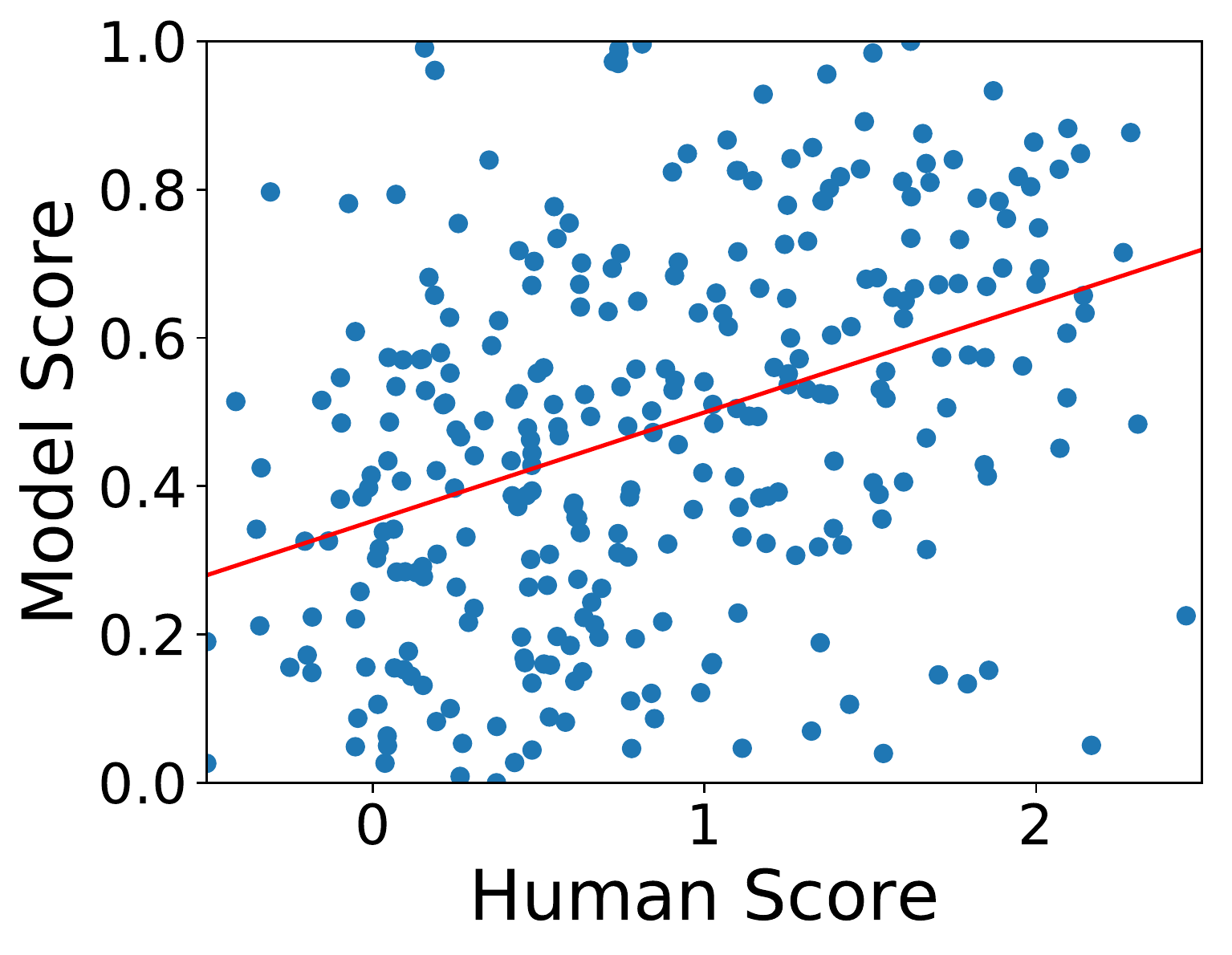} 
	}     
	\subfigure[\ruber\ \scriptsize({Geometric mean})] { \label{fig:top17}     
		\includegraphics[width=0.23\linewidth]{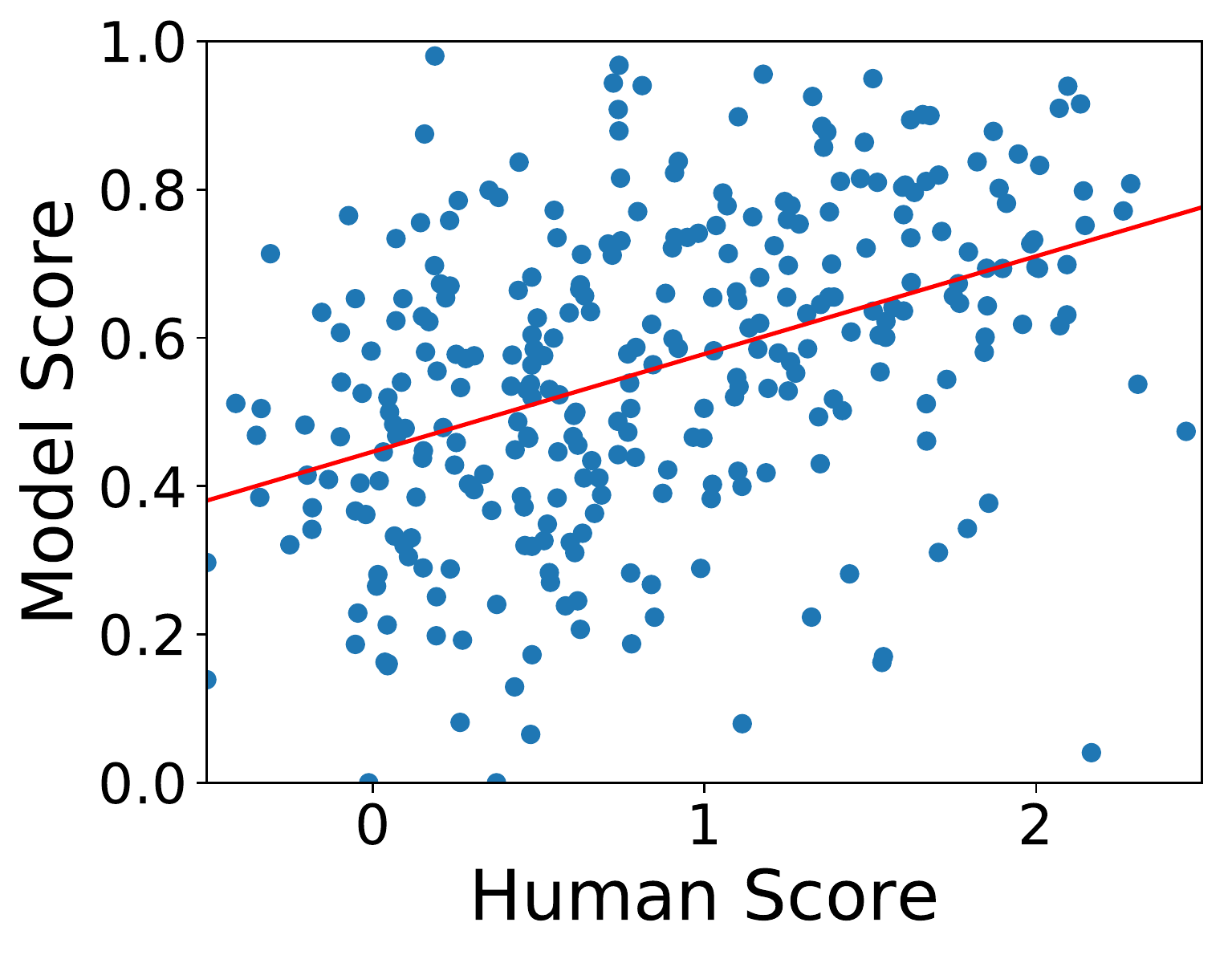} 
	}     
	\subfigure[\ruber\ \scriptsize({Arithmetic mean})] { \label{fig:top18}     
		\includegraphics[width=0.23\linewidth]{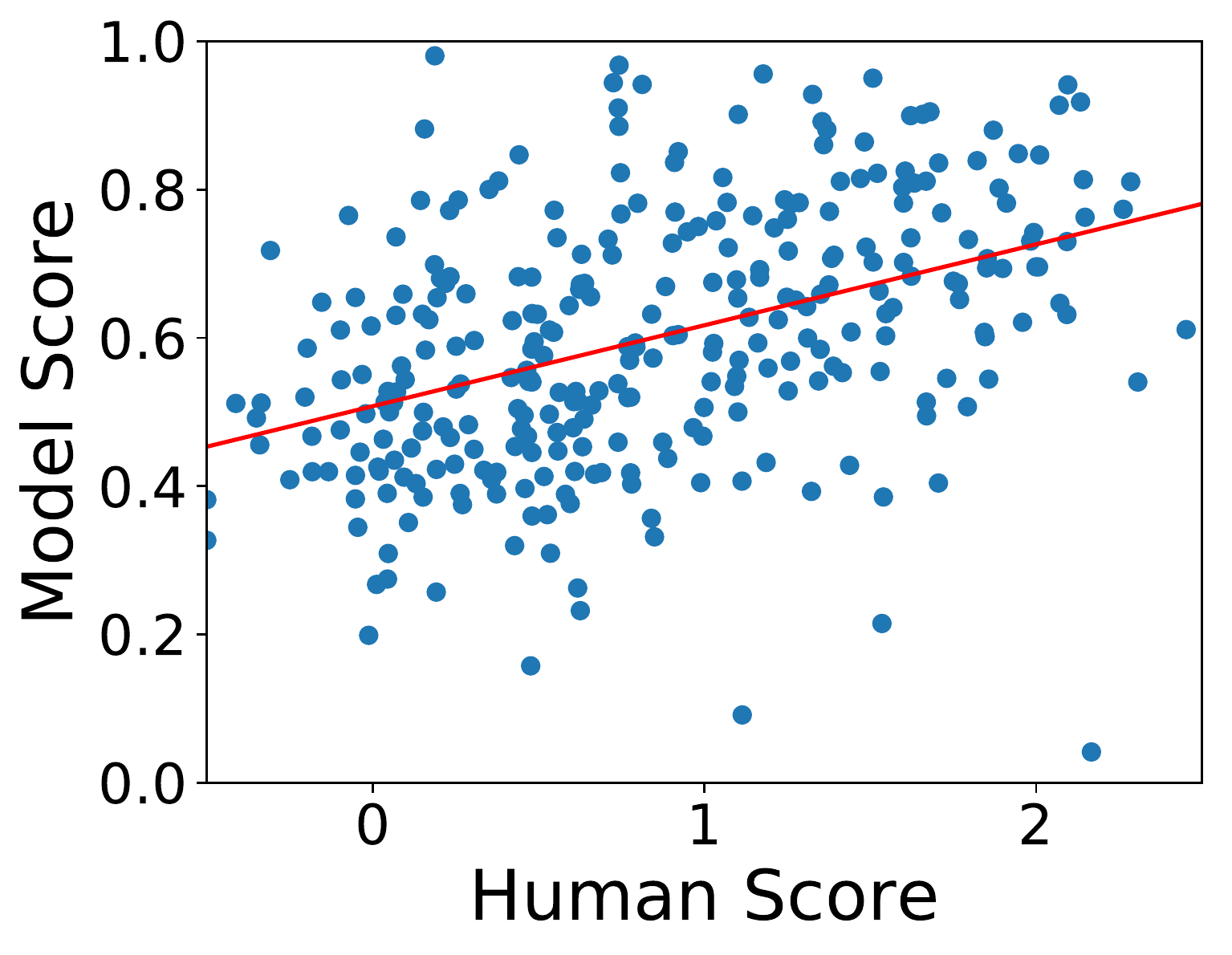} 
	}  
	\caption{Score correlation of the retrieval dialog system. (a) Scatter plot of the medium-correlated human annotator against the rest annotators. (b) Human annotators are divided into two groups, one group vs.~the other. (c)--(h) Scatter plots of different metrics against averaged human scores. Each point is associated with a query-reply pair; we add Guassian noise $\mathcal{N}(0,0.25^2)$ to human scores for a better visualization of point density.}  \label{fig:top1-scatter}
	
	\setlength{\fboxrule}{1pt} 
	\subfigure[Human \scriptsize(1 vs. rest)] { \label{fig:gen12}     
		\includegraphics[width=0.23\linewidth]{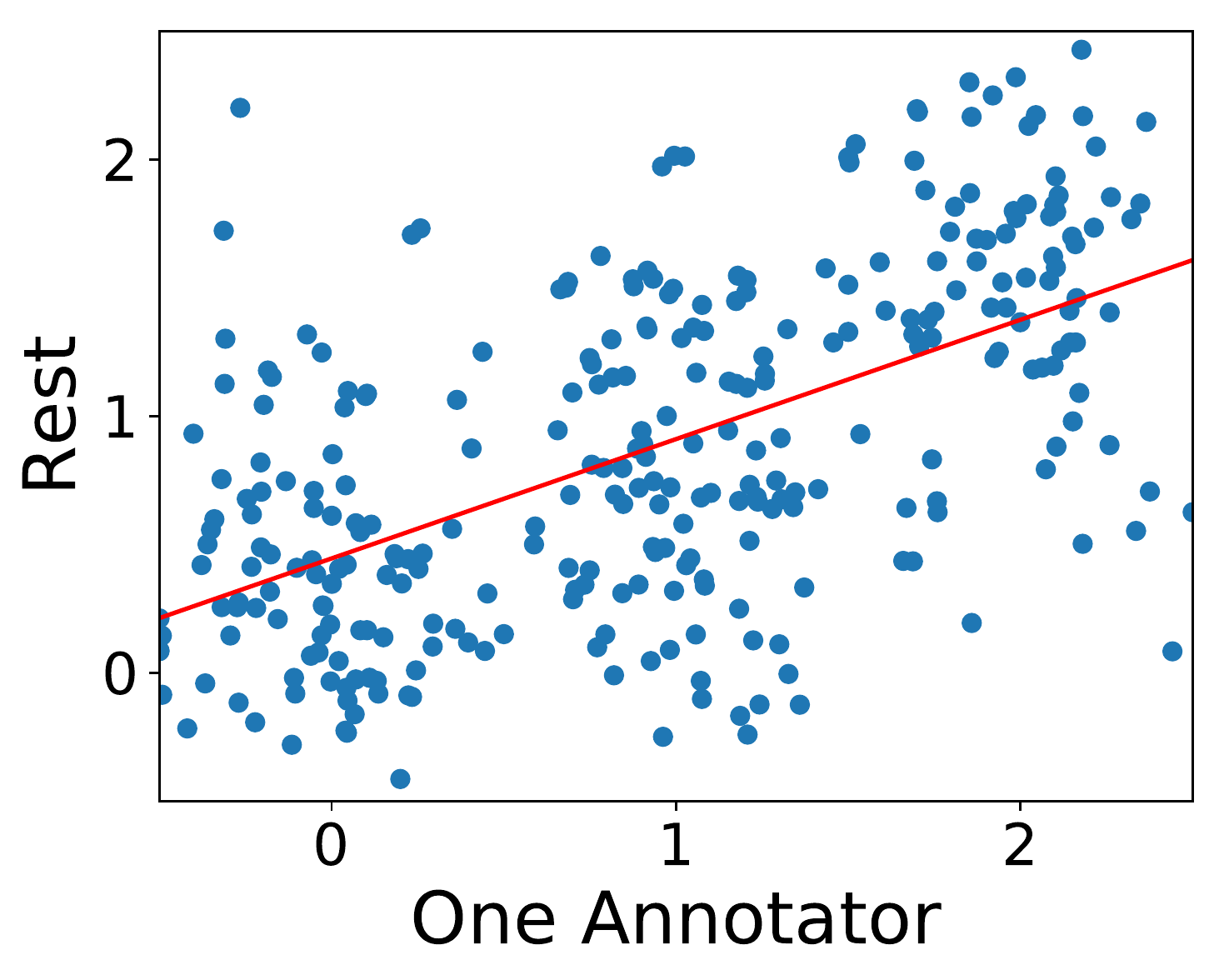} 
	}
	\subfigure[Human \scriptsize(Group 1 vs. Group 2)] { \label{fig:gen11}     
		\includegraphics[width=0.23\linewidth]{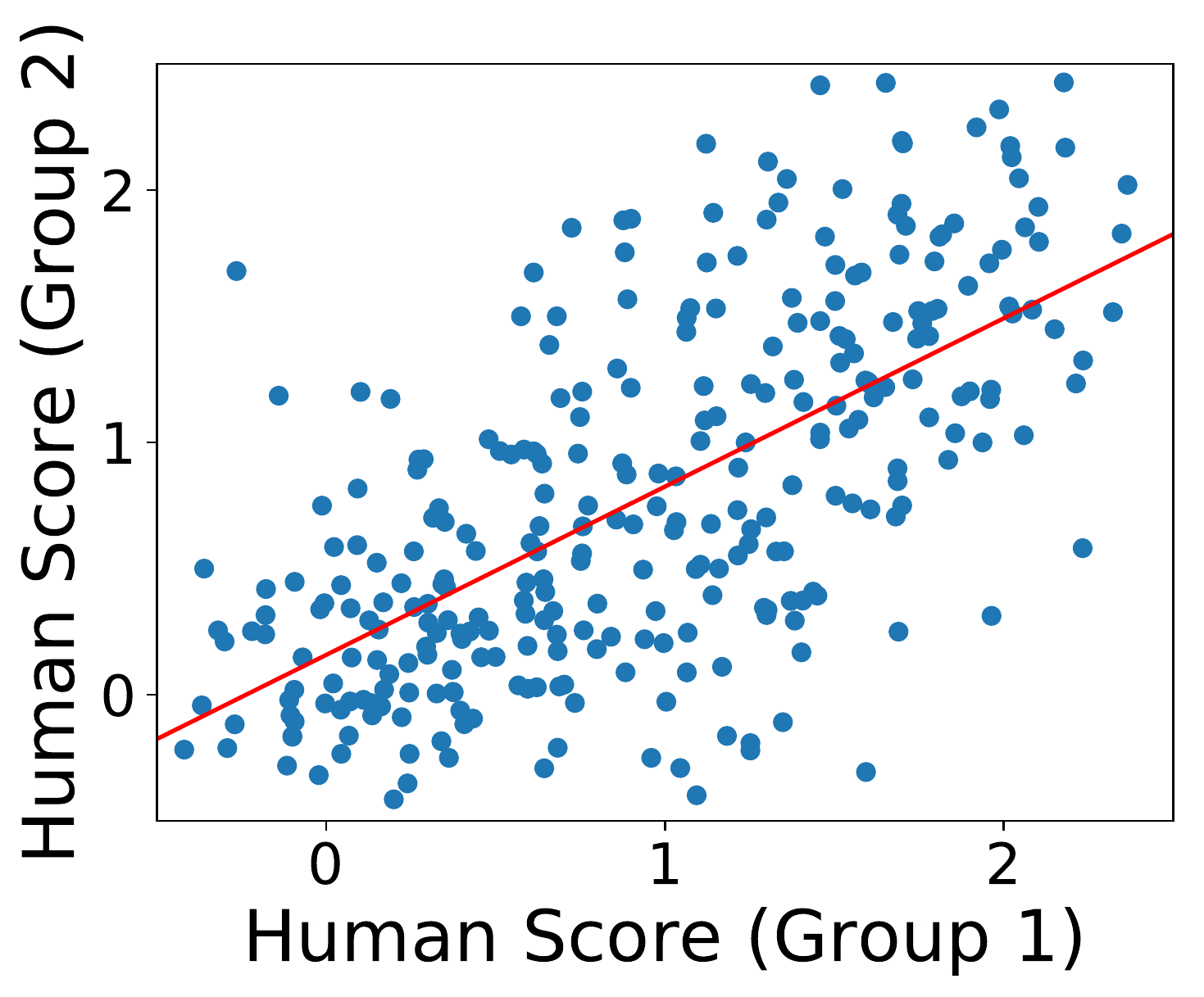} 
	}         
	\subfigure[\bleu-2] { \label{fig:gen13}     
		\includegraphics[width=0.23\linewidth]{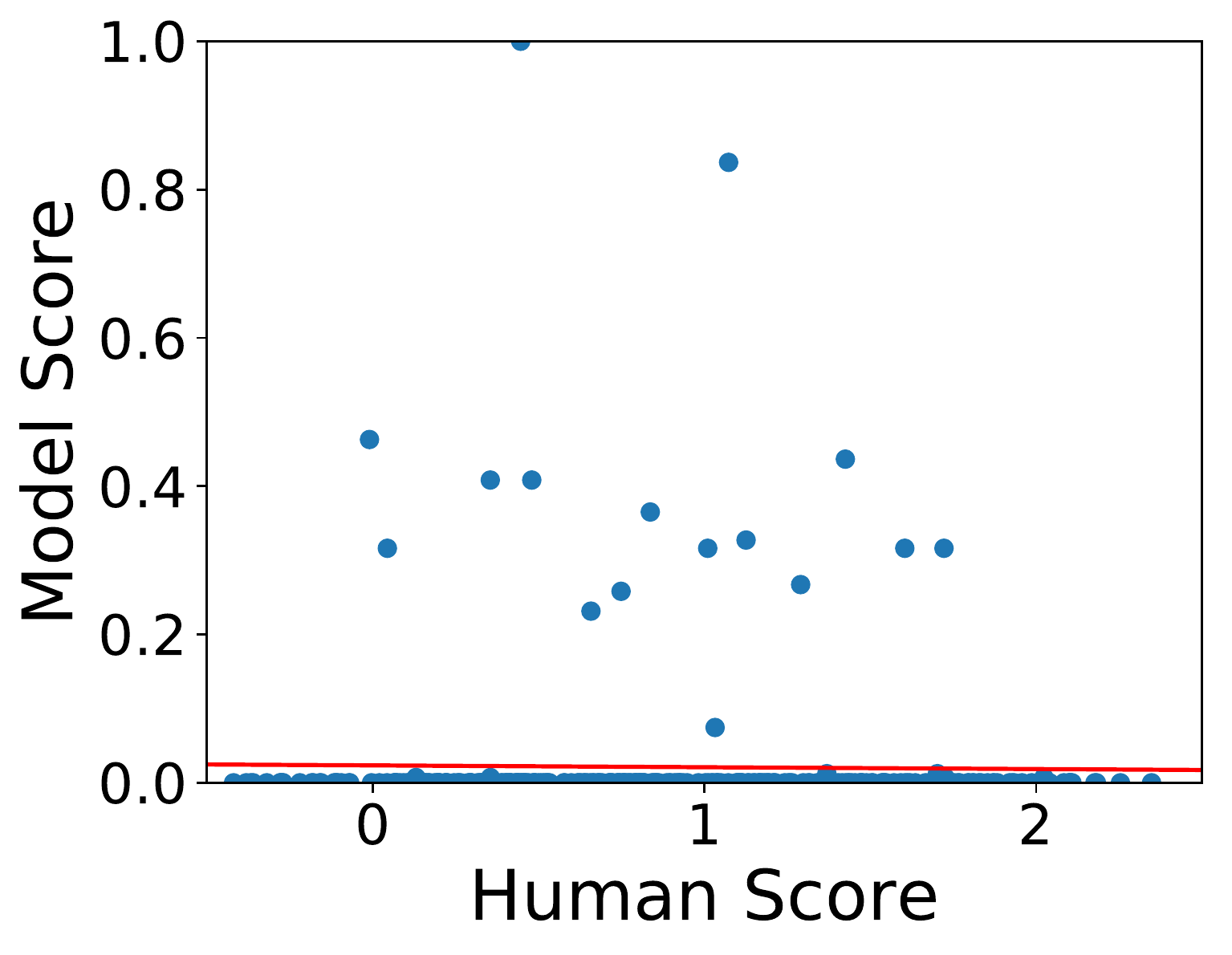} 
	}     
	\subfigure[\rouge] { \label{fig:gen14}     
		\includegraphics[width=0.23\linewidth]{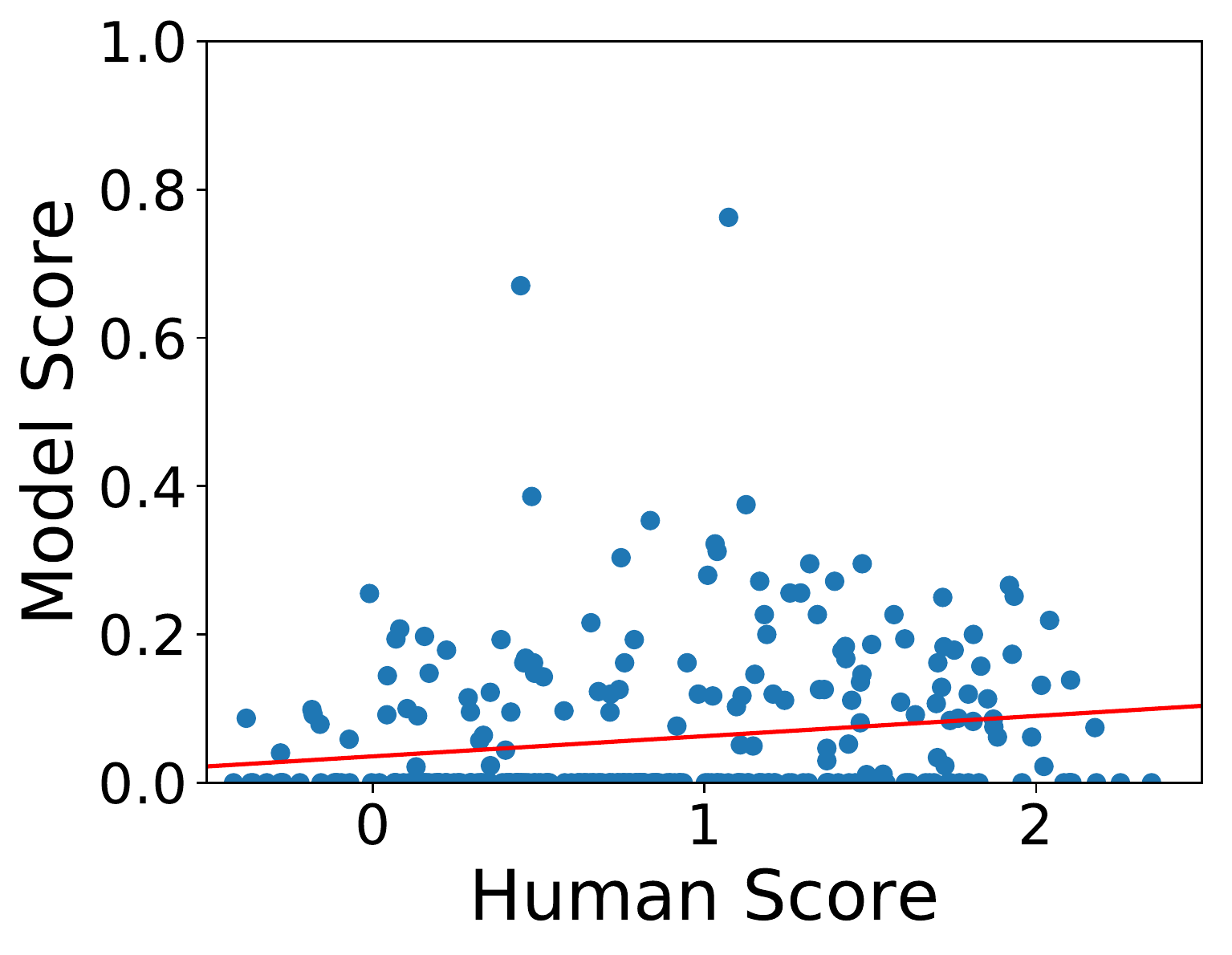} 
	}
	\subfigure[$s_R$ (vector pool)] { \label{fig:gen15}     
		\includegraphics[width=0.23\linewidth]{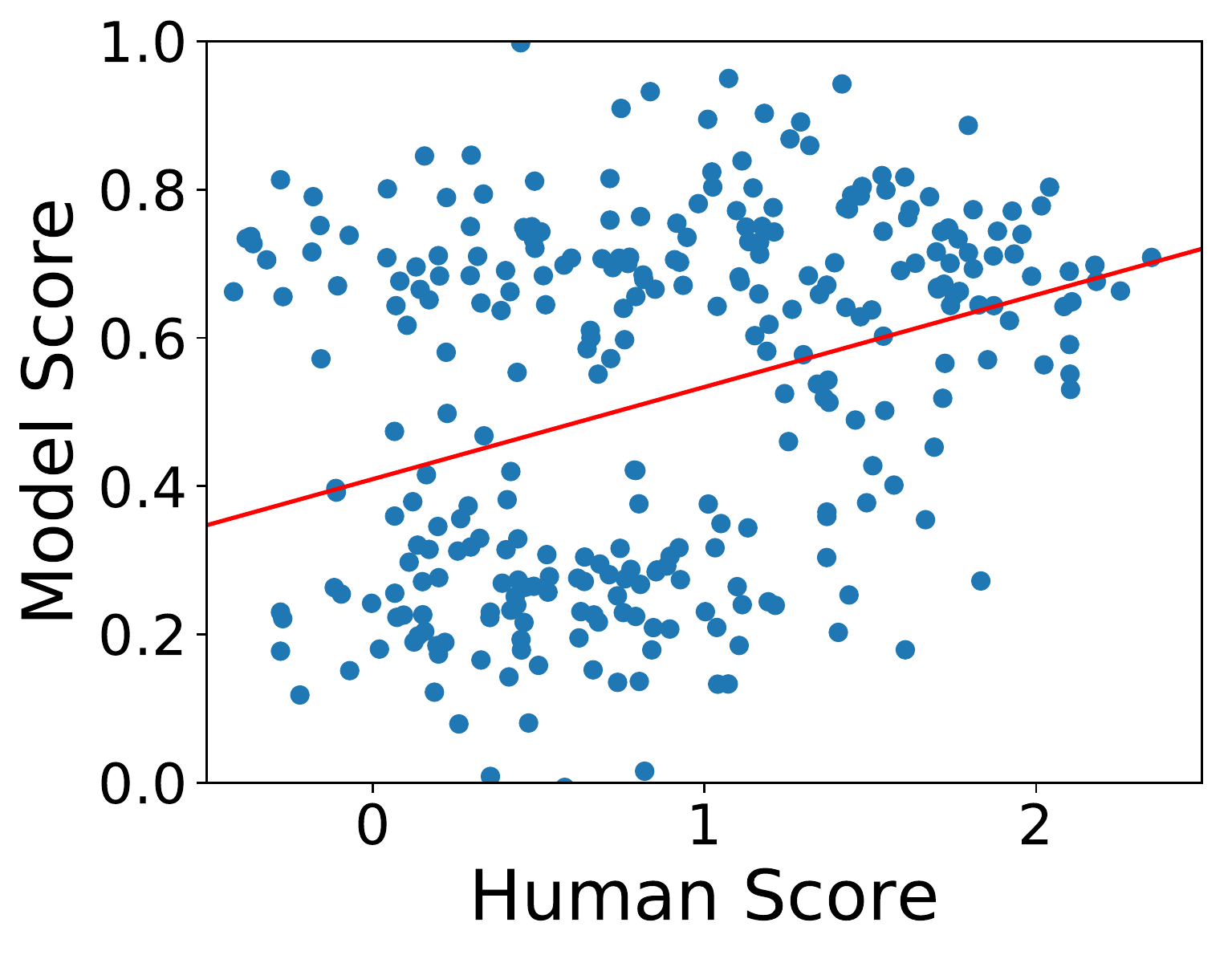} 
	}  
	\subfigure[$s_U$ (NN scorer)] { \label{fig:gen16}     
		\includegraphics[width=0.23\linewidth]{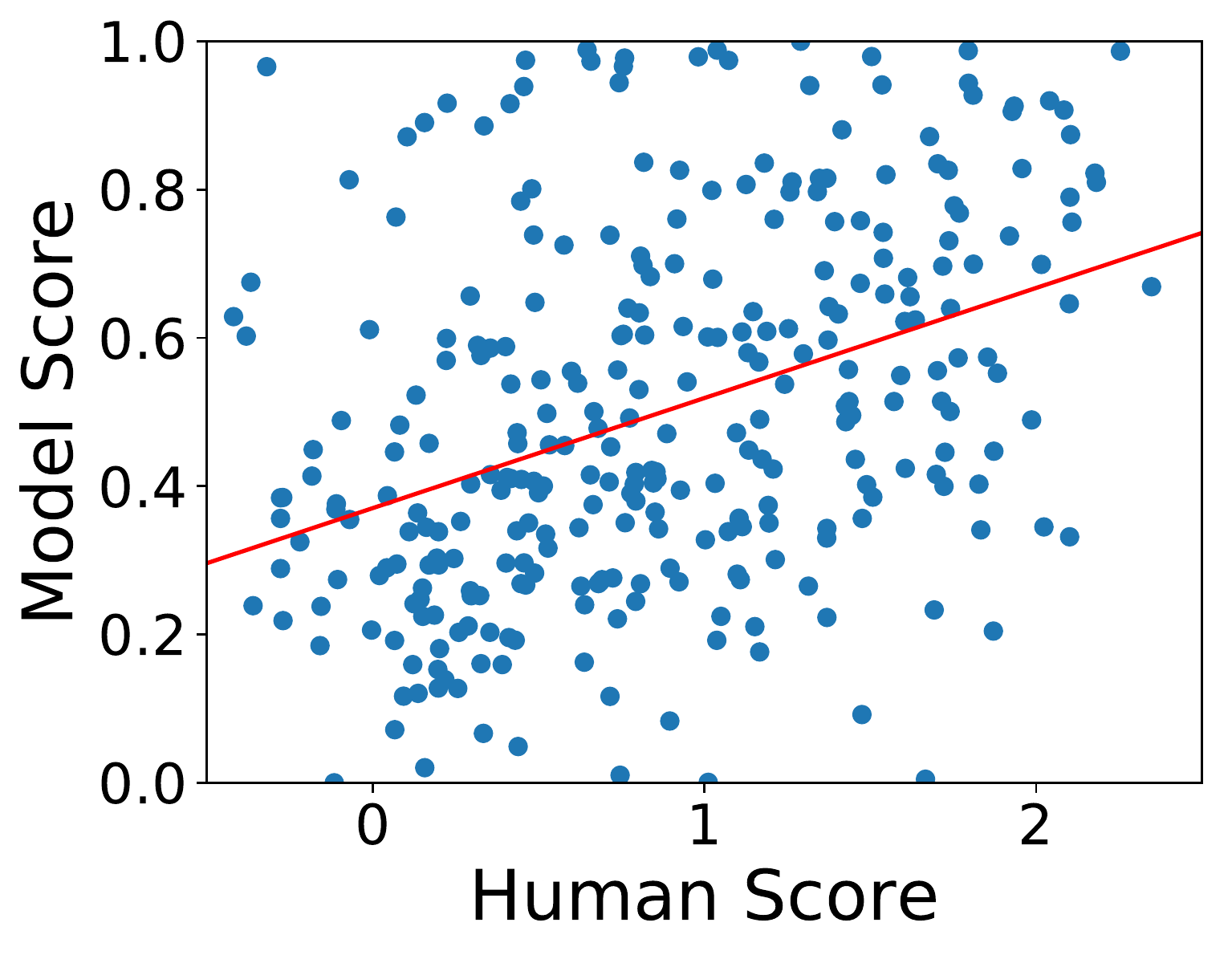} 
	}     
	\subfigure[\ruber\ \scriptsize({Geometric mean})] { \label{fig:gen17}     
		\includegraphics[width=0.23\linewidth]{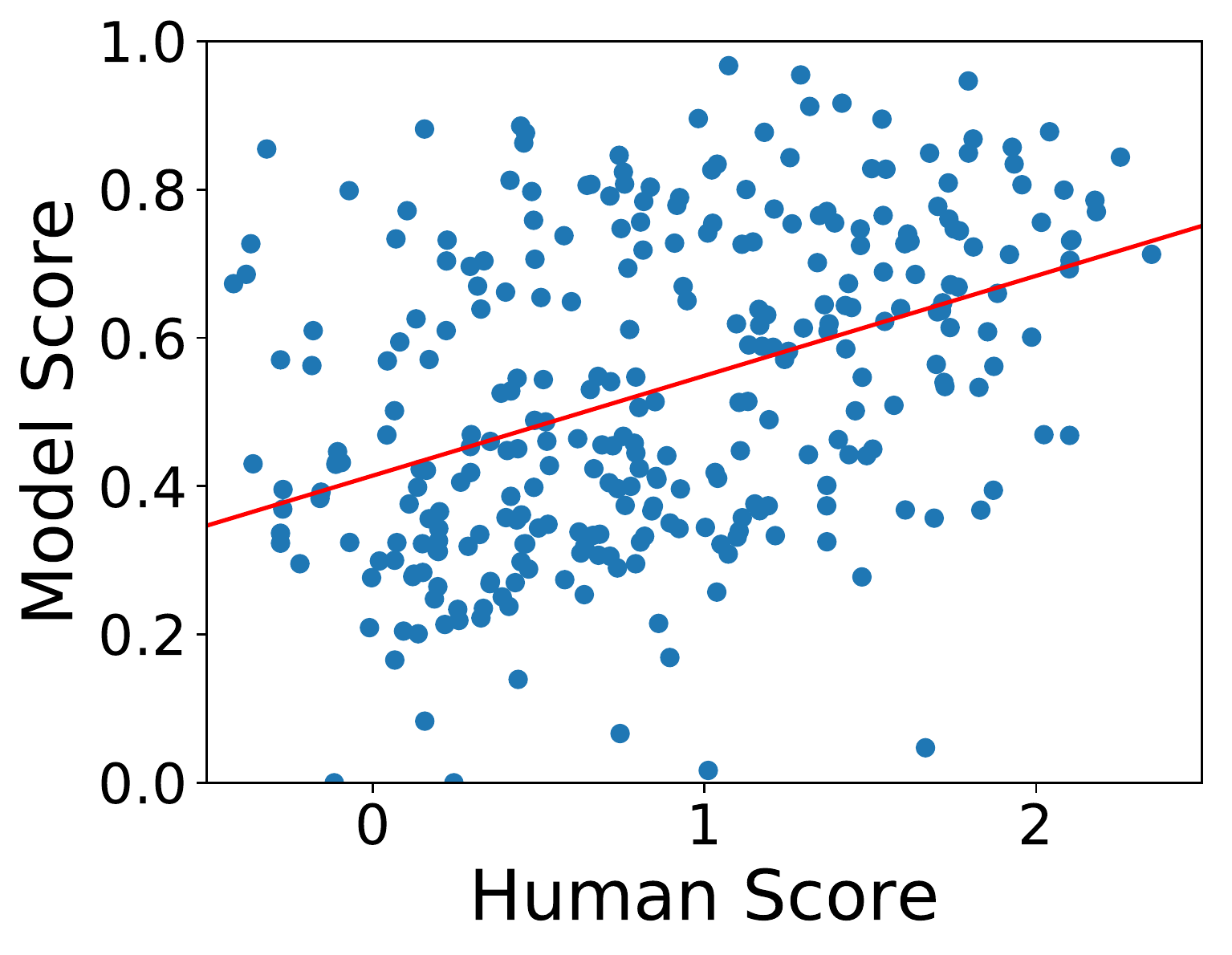} 
	}     
	\subfigure[\ruber\ \scriptsize({Arithmetic mean})] { \label{fig:gen18}     
		\includegraphics[width=0.23\linewidth]{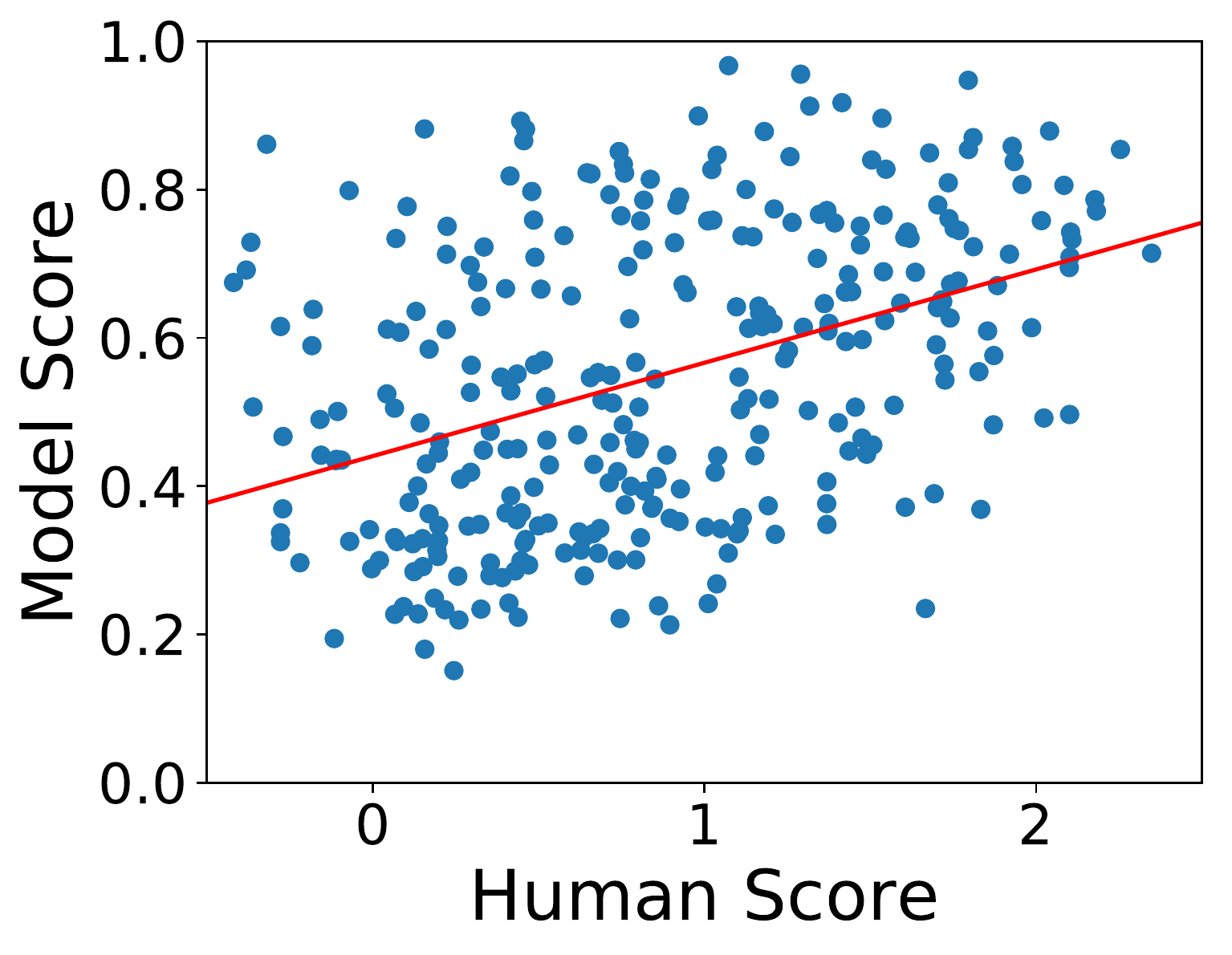} 
	}  
	\caption{Score correlation of the generative dialog system (Seq2Seq w/ attention).} \label{fig:gen1-scatter}     
\end{figure*}

\subsection{Quantitative Analysis}\label{sec:quantitative}
Table~\ref{tab:result} shows the Pearson and Spearman correlation between the proposed \ruber\ metric and human scores; also included are various baselines. Pearson and Spearman correlation are widely used in other research of automatic metrics such as machine translation~\cite{stanojevic2015results}. We compute both correlation based on q/r scores (either obtained or annotated), following~\newcite{liu2016not}.

We find that the referenced metric $s_R$ based on embeddings is more correlated with human annotation than existing metrics including both \bleu\ and \rouge, which are based on word overlapping information. This implies the groundtruth alone is useful for evaluating a candidate reply. But exact word overlapping is too strict in the dialog setting; embedding-based methods measure sentence closeness in a ``soft'' way.

The unreferenced metric $s_U$ achieves even higher correlation than $s_R$, showing that the query alone is also informative and that negative sampling is useful for training evaluation metrics, although it does not require human annotation as labels. 
Our neural network scorer outperforms the embedding-based cosine measure. This is because cosine mainly captures similarity, but the rich semantic relationship between queries and replies necessitates more complicated mechanisms like neural networks.

We combine the referenced and unreferenced metrics as the ultimate \ruber\ approach. Experiments show that choosing the larger value of $s_R$ and $s_U$ (denoted as $\max$) is too lenient, and is slightly worse than other strategies.
Choosing the smaller value ($\min$) and averaging (either geometric or arithmetic mean) yield similar results. While the peak performance is not consistent in two experiments, they significantly outperforms both single metrics, showing the rationale of using a hybrid metric for open-domain dialog systems.
We further notice that our \ruber\ metric has near-human correlation.  More importantly, all components in \ruber\ are heuristic or unsupervised. Thus, \ruber\ does not requirer human labels; it is more flexible than the existing supervised metric~\cite{lowe2017towards}, and can be easily adapted to different datasets.

\subsection{Qualitative Analysis}\label{sec:qualitative}
Figure~\ref{fig:top1-scatter} further illustrates the scatter plots against human judgments for the retrieval system, and Figure~\ref{fig:gen1-scatter} for the generative system (Seq2Seq w/ attention). The two experiments yield similar results and show consistent evidence.

As seen, \bleu\ and \rouge\ scores are zero for most replies, because for short-text conversation extract word overlapping occurs very occasionally; thus these metrics are too sparse.
By contrast, both the referenced and unreferenced scores are not centered at a particular value, and hence are better metrics to use in open-domain dialog systems. Combining these two metrics results in a higher correlation
(Subplots~\ref{fig:top1-scatter}a and~\ref{fig:gen1-scatter}a).

\begin{table*}[!t] 
	\centering
	\resizebox{\linewidth}{!}{
		\begin{tabular}{l|l|l|c|c|c|c|c|c}
			\toprule[2pt]
			\hline
			\multicolumn{1}{c|}{Query} & \multicolumn{1}{c|}{Groundtruth Reply} & \multicolumn{1}{c|}{Candidate Replies} & Human Score & \bleu-2 & \rouge\ & $s_U$ & $s_R$ & \ruber \\
			\hline
			&             &  R1: 我也觉得很近  & \multirow{2}{*} {1.7778}  &  \multirow{2}{*}{0.0000} & \multirow{2}{*}{0.0000}   &   \multirow{2}{*}{1.8867} & \multirow{2}{*}{1.5290}     &  \multirow{2}{*}{1.7078}  \\
			貌似离得挺近的  & 你在哪里的嘞～       & \quad \ \ I also think it's near.  &  &  & &     &       &    \\
			\cline{3-9}
			It seems very near. & Where are you?   & R2: 你哪的？ & \multirow{2}{*}{1.7778} &  \multirow{2}{*}{0.0000} & \multirow{2}{*}{0.7722}  &  \multirow{2}{*}{1.1537} & \multirow{2}{*}{1.7769}  & \multirow{2}{*}{1.4653}   \\
			&             & \quad \ \ Where are you from?  &    &    &   &    &       &     \\
			\hline
			\bottomrule[2pt]       
		\end{tabular}
	}
	\vspace{-4mm}
	\caption{Case study. In the third column, R1 and R2 are obtained by the generative and retrieval systems, resp. \ruber\ here uses arithmetic mean. For comparison, we normalize all scores to  the range of human annotation, i.e., $[0, 2]$. Note that the normalization does not change the degree of correlation. }	\vspace{-4mm}
	\label{tab:casestudy}
\end{table*}

\begin{table}[!t]
	\centering
	\resizebox{\linewidth}{!}{
		\begin{tabular}{c|c|c|c}
			\toprule[2pt] \hline
			\multicolumn{2}{c|}{\multirow{2}[4]{*}{Metrics}} 
			&\multicolumn{2}{c}{\textbf{Seq2Seq (w/ attention)}} \\
			\cline{3-4}    \multicolumn{2}{c|}{}  & Pearson\scriptsize($p$-value)  & Spearman\scriptsize($p$-value)  \\
			\hline \hline
			\multirow{2}[2]{*}{Inter-annotator} 
			& Human (Avg) & 0.4860{\scriptsize($<\!0.01$)} & 0.4890{\scriptsize($<\!0.01$)} \\
			& Human (Max) & 0.6500{\scriptsize($<\!0.01$)} & 0.6302{\scriptsize($<\!0.01$)} \\
			\hline  
			\multirow{6}[2]{*}{Referenced} & \bleu-1 & 0.2091{\scriptsize(0.0102)} & 0.2363{\scriptsize($<\!0.01$)} \\
			& \bleu-2 & 0.0369{\scriptsize(0.6539)} & 0.0715{\scriptsize(0.3849)} \\
			& \bleu-3 & 0.1327{\scriptsize(0.1055)} & 0.1299{\scriptsize(0.1132)} \\
			& \bleu-4 & nan   & nan \\
			& \rouge & 0.2435{\scriptsize($<\!0.01$)} & 0.2404{\scriptsize($<\!0.01$)} \\
			& Vector pool ($s_R$) & 0.2729{\scriptsize($<\!0.01$)} & 0.2487{\scriptsize($<\!0.01$)} \\
			\hline  
			\multirow{2}[2]{*}{Unreferenced} & Vector pool & 0.2690{\scriptsize($<\!0.01$)} & 0.2431{\scriptsize($<\!0.01$)} \\
			& NN scorer ($s_U$)  & 0.2911{\scriptsize($<\!0.01$)} & 0.2562{\scriptsize($<\!0.01$)} \\
			\hline  
			\multirow{4}[2]{*}{\ruber} & Min   & 0.3629{\scriptsize($<\!0.01$)} & 0.3238{\scriptsize($<\!0.01$)} \\
			& Geometric mean & \textbf{0.3885}{\scriptsize($<\!0.01$)} & \textbf{0.3462}{\scriptsize($<\!0.01$)} \\
			& Arithmetic mean & 0.3593{\scriptsize($<\!0.01$)} & 0.3304{\scriptsize($<\!0.01$)} \\
			& Max   & 0.2702{\scriptsize($<\!0.01$)} & 0.2778{\scriptsize($<\!0.01$)} \\
			\hline
			\bottomrule[2pt]
		\end{tabular}
	}	\vspace{-4mm}
	\caption{Correlation between automatic metrics and human annotation in the transfer setting.}
		\vspace{-4mm}
	\label{tab:result2}
\end{table}

We would like to clarify more regarding human-human plots. \newcite{liu2016not} group human annotators into two groups and show scatter plots between the two groups, the results of which in our experiments are shown in Subplots~\ref{fig:top1-scatter}b and~\ref{fig:gen1-scatter}b. However, in such plots, each data point's score is averaged over several annotators, resulting in low variance of the value. It is not a right statistic to compare with.\footnote{In the limit of the annotator number to infinity, Subplots~\ref{fig:top1-scatter}b and~\ref{fig:gen1-scatter}b would become diagonals (due to the Law of Large Numbers).}
In our experimental design, we would like to show the difference between a single human annotator versus the rest annotators; in particular, the scatter plots \ref{fig:top1-scatter}a and~\ref{fig:gen1-scatter}a demonstrate the median-correlated human's performance. These qualitative results show our \ruber\ metric achieves similar correlation to humans.

\subsection{Case Study}
Table~\ref{tab:casestudy} illustrates an example of our metrics as well as baselines. We see that \bleu\ and \rouge\ scores are prone to being zero. Even the second reply is very similar to the groundtruth, its Chinese utterances do not have bi-gram overlap, resulting in a \bleu-2 score of zero. By contrast, our referenced and unreferenced metrics are denser and more suited to open-domain dialog systems.

We further observe that the referenced metric $s_R$ assigns a high score to R1 due to its correlation with the query, whereas the unreferenced metric $s_U$ assigns a high score to R2 as it closely resembles the groundtruth. Both R1 and R2 are considered reasonable by most annotators, and our \ruber\ metric yields similar scores to human annotation by balancing $s_U$ and $s_R$.

\newcommand{\minitab}[2][c]{\begin{tabular}{#1}#2\end{tabular}}

\subsection{Transferability}
We would like to see if the \ruber\ metric can be transferred to different datasets. Moreover, we hope \ruber\ can be directly adapted to other datasets even without re-training the parameters. 

We crawled another Chinese dialog corpus from the Baidu Tieba\footnote{\url{http://tieba.baidu.com}} forum. 
The dataset comprises 480k query-reply pairs, and its topics may vary from the previously used Douban corpus. We only evaluated the results of the Seq2Seq model (with attention) because of the limit of time and space.

\begin{figure}[!t] \centering
	\subfigure[Human \scriptsize(1 vs. rest)] { \label{fig:tans12}     
		\includegraphics[width=0.45\linewidth]{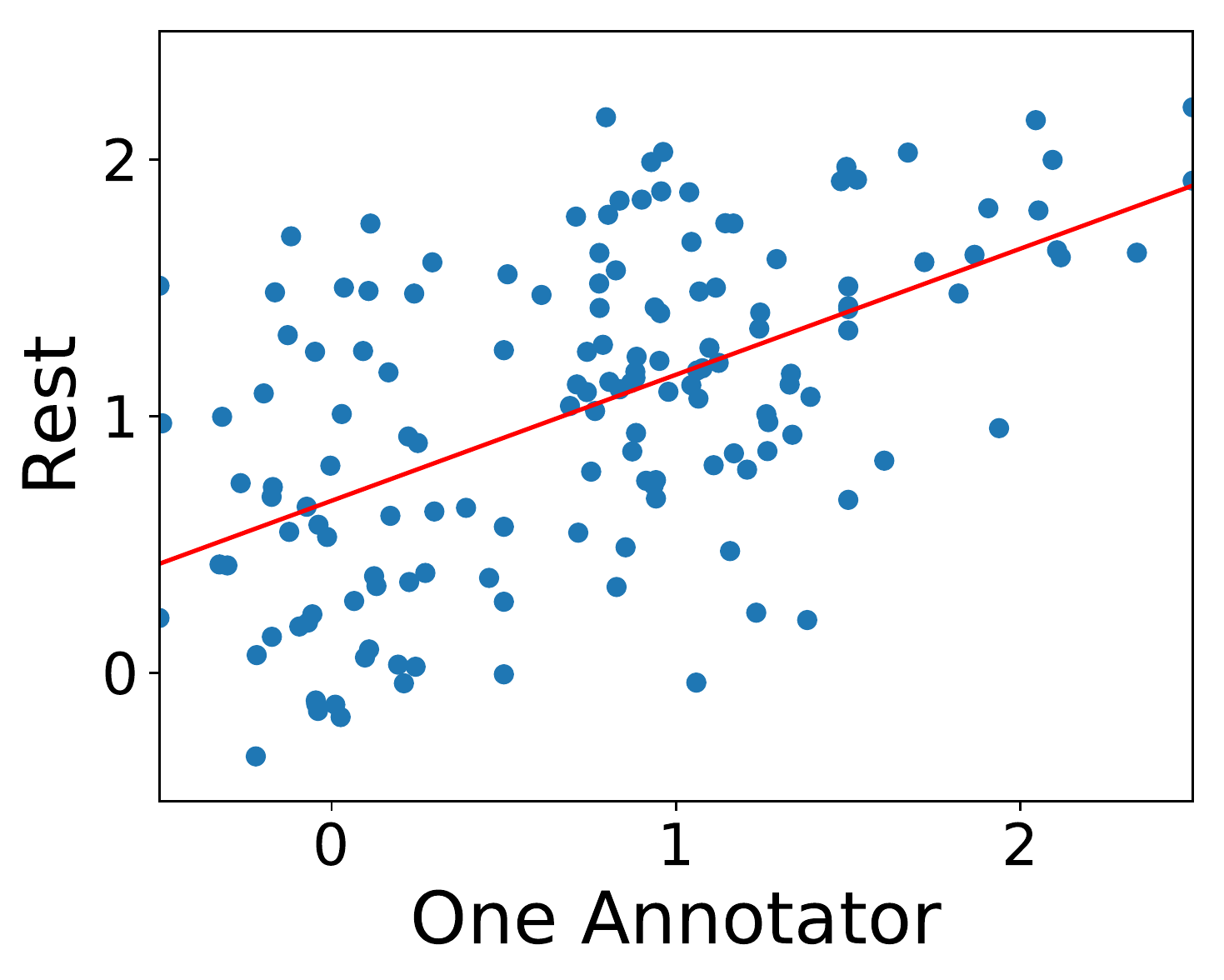} 
	}   
	\subfigure[\bleu-2] { \label{fig:tans13}     
		\includegraphics[width=0.45\linewidth]{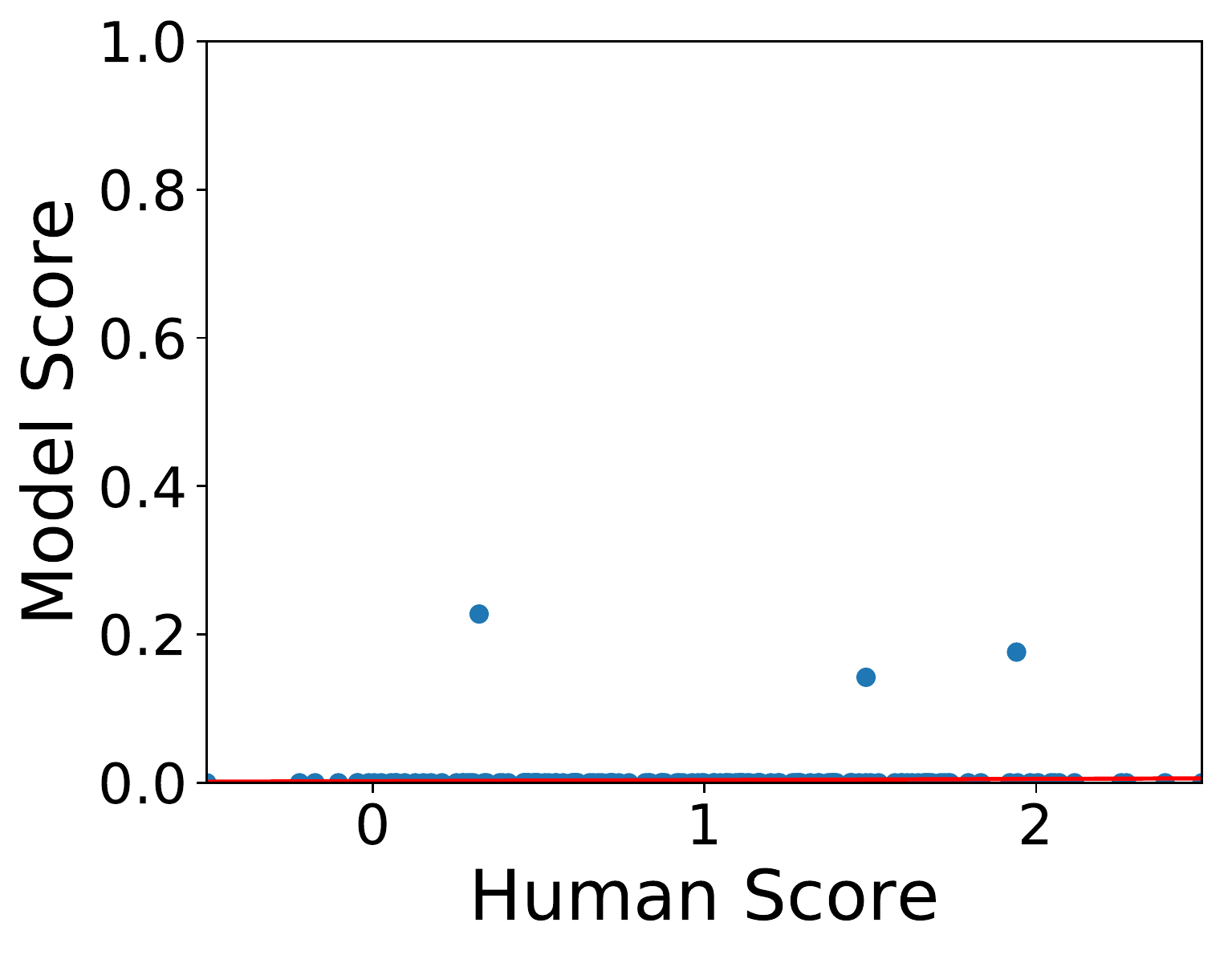} 
	}     
	\subfigure[\rouge] { \label{fig:tans14}     
		\includegraphics[width=0.45\linewidth]{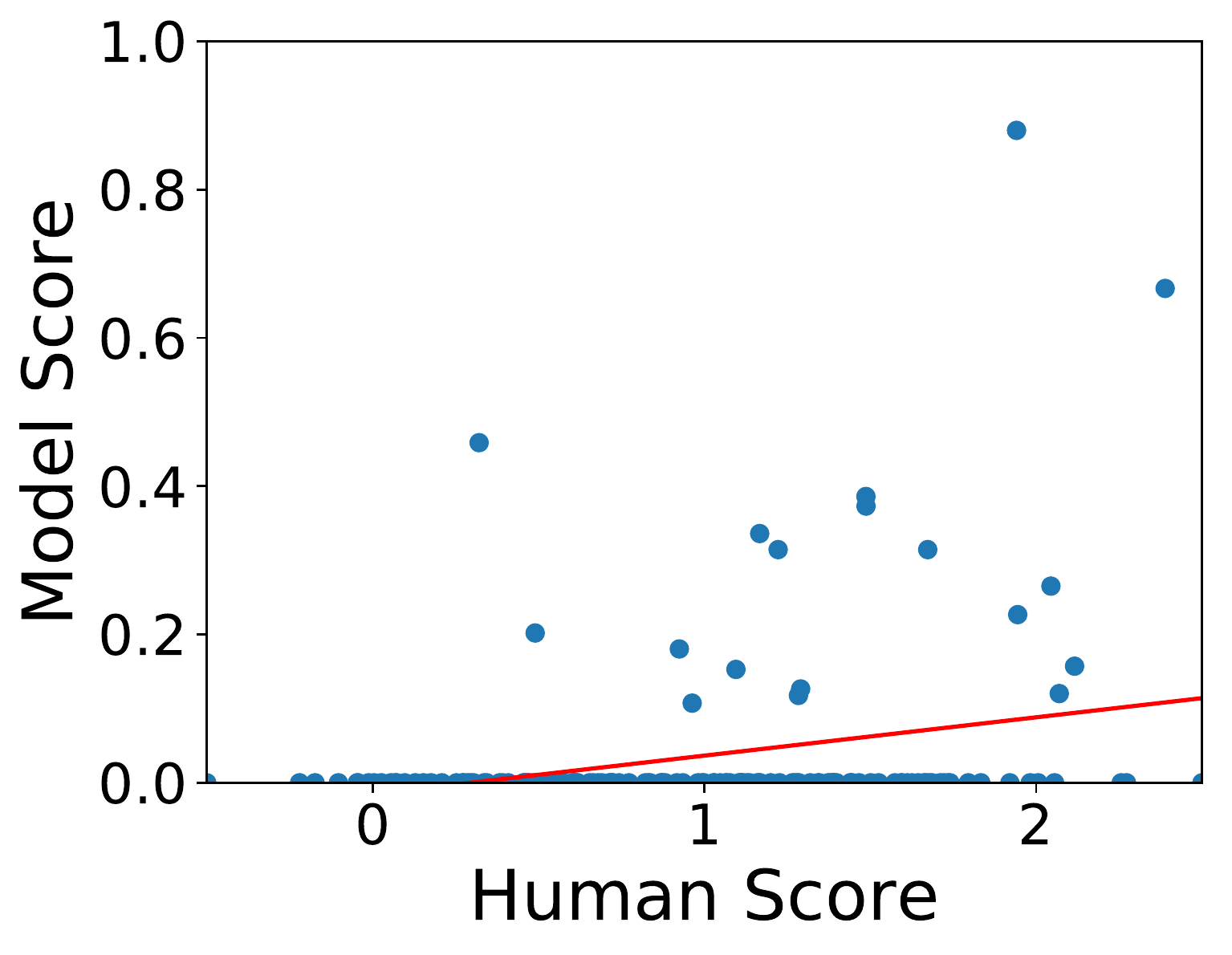} 
	}
	\subfigure[\ruber\ \scriptsize({Geometric mean})] { \label{fig:tans17}     
		\includegraphics[width=0.45\linewidth]{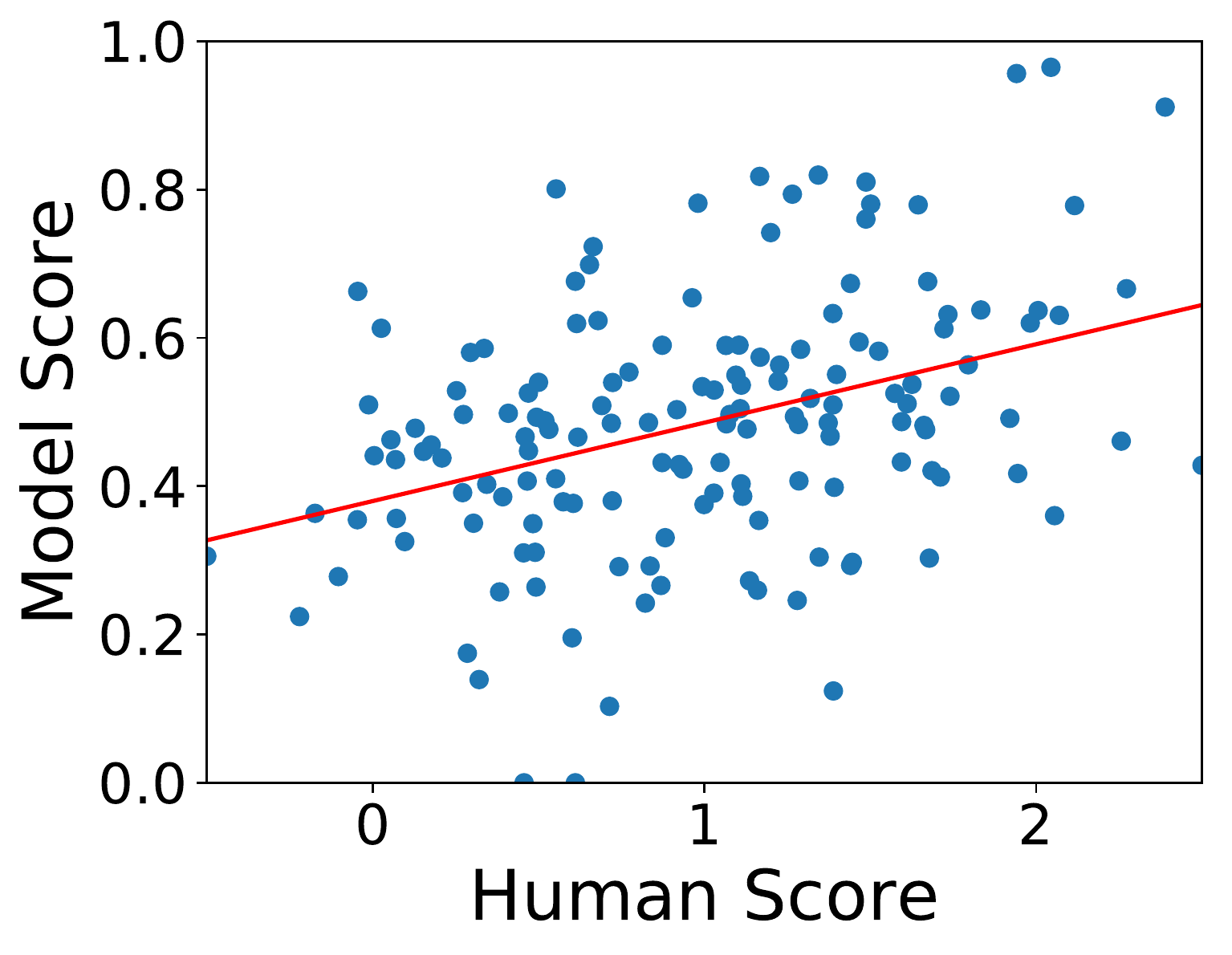} 
	}     
	\vspace{-4mm}
	\caption{Score correlation of the generative dialog system (Seq2Seq w/ attention) in the transfer setting.}  
	\label{fig:trans-scatter}
\end{figure}
We directly applied the \ruber\ metric to the Baidu dataset, i.e., word embeddings and $s_R$'s parameters were trained on the Douban dataset. We also had 9 volunteers to annotate 150 query-reply pairs, as described in Section~\ref{sec:set}. Table~\ref{tab:result2} shows the Pearson and Spearman correlation and Figure~\ref{fig:trans-scatter} demonstrates the scatter plots in the transfer setting.

As we see, transferring to different datasets leads to slight performance degradation compared with Table~\ref{tab:result}.
This makes sense because the parameters, especially the $s_R$ scorer's, are not trained for the Tieba dataset.
That being said, \ruber\ still significantly outperforms baseline metrics, showing fair transferability of our proposed method.

Regarding different blending methods, min and geometric/arithmetic mean are similar and better than the max operator; they also outperform their components $s_R$ and $s_U$. The results are consistent with the non-transfer setting (Subsections~\ref{sec:quantitative} and~\ref{sec:qualitative}), showing additional evidence of the effectiveness of our hybrid approach.

\section{Related Work}
\subsection{Automatic Evaluation Metrics} 
Automatic evaluation is crucial to the research of language generation tasks such as dialog systems~\cite{li2015diversity}, machine translation~\cite{papineni2002bleu}, and text summarization~\cite{rouge}. The Workshop on Machine Translation (WMT) organizes shared tasks for evaluation metrics~\cite{stanojevic2015results,bojar2016results}, attracting a large number of researchers and greatly promoting the development of translation models. 

Most existing metrics evaluate generated sentences by word overlapping against a groundtruth sentence. For example, \textsc{Bleu}~\cite{papineni2002bleu} computes geometric mean of the precision for $n$-gram ($n=1,\cdots,4$); NIST~\cite{doddington2002automatic} replaces geometric mean with arithmatic mean. Summarization tasks prefer recall-oriented metrics like \rouge~\cite{rouge}. \textsc{Meteor}~\cite{banerjee2005meteor} considers precision as well as recall for more comprehensive matching.
Besides, several metrics explore the source information to evaluate the target without referring to the groundtruth. \newcite{popovic2011evaluation} evaluate the translation quality by calculating the probability score based on IBM Model \uppercase\expandafter{\romannumeral1} between words in the source and target sentences. \newcite{louis2013automatically} use the distribution similarity between input and generated summaries to evaluate the quality of summary content. 

From the machine learning perspective, automatic evaluation metrics can be divided into non-learnable and learnable approaches. Non-learnable metrics (e.g., \bleu\ and \rouge) typically measure the quality of generated sentences by heuristics (manually defined equations), whereas learnable metrics are built on machine learning techniques. 
\newcite{specia2010machine} and \newcite{avramidis2011evaluate} train a classifier to judgment the quality with linguistic features extracted from the source sentence and its translation. Other studies regard machine translation evaluation as a regression task supervised by manually annotated scores~\cite{albrecht2007regression,gim2008heterogeneous,specia2009estimating}.

Compared with traditional heuristic evaluation metrics, learnable metrics can integrate linguistic features\footnote{Technically speaking, existing metrics (e.g., \bleu\ and \textsc{Meteor}) can be regarded as features extracted from the output sentence and the groundtruth.} to enhance the correlation with human judgments through supervised learning. However, handcrafted features often require expensive human labor, but do not generalize well. More importantly, these learnable metrics require massive human-annotated scores to learn the model parameters. Different from the above methods, our proposed metric apply negative sampling to train the neural network to measure the relatedness of query-reply pairs, and thus can extract features automatically without any supervision of human-annotated scores.

\subsection{Evaluation for Dialog Systems}
Dialog systems based on generative methods are also language generation tasks, and thus several researchers adopt \bleu\  score to measure the quality of a reply~\cite{li2015diversity,sordoni2015neural,song2016two}. However, its effectiveness has been questioned~\cite{callison2006re,galley2015deltableu}. Meanwhile, \newcite{liu2016not}  conduct extensive empirical experiments and show the weak correlation of existing metrics (e.g., \bleu, \rouge\ and \textsc{meteor}) with human judgements for dialog systems. Based on \bleu, \newcite{galley2015deltableu} propose $\Delta$\textsc{Bleu}, which considers several reference replies. However, multiple references are hard to obtain in practice.

Recent advances in generative dialog systems have raised the problem of universally relevant replies. \newcite{li2015diversity} measure the reply diversity by calculating the proportion of distinct unigrams and bigrams. Besides, \newcite{serban2016hierarchical} and \newcite{mou2016sequence} use entropy to measure the information of generated replies, but such metric is independent of the query and groundtruth. Compared with the neural network-based metric proposed by \newcite{lowe2017towards}, our approach does not require human-annotated scores.

\section{Conclusion and Discussion}
In this paper, we proposed an evaluation methodology for open-domain dialog systems. Our metric is called \ruber\ (a \textit{Referenced metric and Unreferenced metric Blended Evaluation Routine}), as it considers both the groundtruth and its query.
Experiments show that, although unsupervised, \ruber\ has strong correlation with human annotation, and has fair transferability over different open-domain datasets.

Our paper currently focuses on single-turn conversation as a starting point of our research. However, the \ruber\ framework can be extended naturally to more complicated scenarios: in a history/context-aware dialog system, for example, the modification shall lie in designing the neural network, which will take context into account, for the unreferenced metric.

\bibliographystyle{acl_natbib}
\bibliography{ruber}

\end{CJK*}
\end{document}